\documentclass{article}
\usepackage[preprint]{colm2026_conference}

\usepackage{hyperref}
\usepackage{algorithm}
\usepackage{algorithmic}
\usepackage{microtype}
\usepackage{lineno}

\definecolor{darkblue}{rgb}{0, 0, 0.5}
\hypersetup{colorlinks=true, citecolor=darkblue, linkcolor=darkblue, urlcolor=darkblue}

\usepackage{amsmath,amsfonts,bm}

\def\eqref#1{equation~\ref{#1}}

\def\1{\bm{1}}

\DeclareMathAlphabet{\mathsfit}{\encodingdefault}{\sfdefault}{m}{sl}
\SetMathAlphabet{\mathsfit}{bold}{\encodingdefault}{\sfdefault}{bx}{n}

\usepackage{url}
\usepackage{booktabs}
\usepackage{amsfonts}
\usepackage{nicefrac}
\usepackage{xcolor}
\usepackage{graphicx}
\usepackage{multirow}
\usepackage{array}
\usepackage[table]{xcolor}
\usepackage{tcolorbox}
\usepackage{enumitem}
\usepackage{listings}
\usepackage[T1]{fontenc}
\usepackage{colortbl}
\usepackage{subcaption}
\usepackage{wrapfig}
\usepackage{adjustbox}
\usepackage{tcolorbox}
\tcbuselibrary{breakable,skins}
\usepackage{caption}

\newcommand{\optcolor}[1]{%
  \ifdim #1pt > 50pt \cellcolor{green!20}{#1}%
  \else\ifdim #1pt > 30pt \cellcolor{orange!20}{#1}%
  \else\ifdim #1pt > 20pt \cellcolor{red!10}{#1}%
  \else \cellcolor{red!20}{#1}%
  \fi\fi\fi
}
\newcommand{\errcolor}[1]{%
  \ifdim #1pt > 8pt \cellcolor{red!35}{#1}%
  \else\ifdim #1pt > 3pt \cellcolor{orange!25}{#1}%
  \else\ifdim #1pt > 1pt \cellcolor{yellow!20}{#1}%
  \else \cellcolor{green!15}{#1}%
  \fi\fi\fi
}

\newcommand{\draftonly}[1]{#1}
\newcommand{\draftcomment}[1]{\draftonly{#1}}

\newcommand{\sq}[1]{\draftcomment{\textcolor{orange}{\small [SQ: #1]}}}

\title{\textit{COMPASS}:\\ Benchmarking Constrained Optimization in LLM Agents}

\author{
Tian Qin\thanks{Work done during internship at Apple. Correspondence to \texttt{tqin@g.harvard.edu}.} \\
Harvard University
\And
Haoping Bai \\
Apple
\And
Ting-Yao Hu \\
Apple
\And
Raviteja Vemulapalli \\
Apple
\AND
Hema Swetha Koppula \\
Apple
\And
Zhiyang Xu\footnotemark[1] \\
Virginia Tech
\And
Bowen Jin\footnotemark[1] \\
UIUC
\And
Mert Cemri\footnotemark[1] \\
UC Berkeley
\AND
Jiarui Lu \\
Apple
\And
Zirui Wang \\
Apple
\And
Meng Cao \\
Apple
}

\begin{document}

\ifcolmsubmission
\linenumbers
\fi

\maketitle

\begin{abstract}
    Human decision-making often involves constrained optimization. As LLM agents are deployed to assist with real-world tasks like travel planning, shopping, and scheduling, they must mirror this capability. We introduce \textit{COMPASS}, a benchmark that evaluates whether LLM agents can perform constrained optimization in realistic travel planning settings. 
    To success in these tasks, agents must engage in multi-turn conversations with user to gather task information as well as use tools to gather information from the database. Then agents must propose a solution that not only satisfies hard constraints but also optimizes user's utility objective. Evaluating state-of-the-art models, we reveal a significant \textbf{feasible-optimal gap}: while models achieve 70-90\% feasibility (constraint satisfaction), they reach only 20-60\% optimality (utility optimization). Our analysis shows that tool use is \textit{not} the bottleneck. Instead, the core limitation is insufficient exploration of the search space, with success strongly correlating with information gathered. Coding agents show a promising approach to mitigate this gap. Together, \textit{COMPASS} provides a testbed for developing LLM agents that can truly mirror human decision-making by both satisfying constraints and optimizing objectives.\looseness=-1
\end{abstract}

\section{Introduction}
\label{intro}
In economics, operations research, and decision theory, humans are modeled as decision makers who are constantly maximizing their utility subject to constraints. When selecting products or services, decision-makers naturally partition their preferences into two categories: \textit{hard constraints} define the feasible set of options (e.g., ``\textit{I cannot exceed \$500}" or ``\textit{I must arrive by Friday}"), and \textit{utility objective function} rank options within this feasible set (e.g., "\textit{I prefer shorter travel time}" or "\textit{I value higher ratings}"). As Large Language Models (LLMs) are deployed to assist users with everyday tasks such as travel planning, shopping, scheduling, they must mirror this capability.

When deployed in the real world, constrained optimization is more than a mathematical problem. Agents must actively use tools to discover and compare options, and learn the user's hard constraints and utility function through multi-turn interaction. Yet existing benchmarks address only pieces of this challenge: travel planning benchmarks focus mostly on constraint satisfaction~\citep{Xie2024-gx, Kohli2024-hl}, while tool-use benchmarks test only whether agents call correct tools with valid parameters~\citep{Zhong2025-yi, Yao2024-aa}. Neither line of work tests whether agents can \textit{strategically} explore a search space to discover optimal solutions. Travel planning is our chosen instantiation of this problem, but the underlying challenge is general: any domain with combinatorial structure, and user-specified constraints and preferences poses the same challenge on an agent.

\begin{figure*}[t]
  \centering
  \includegraphics[width=0.9\linewidth]{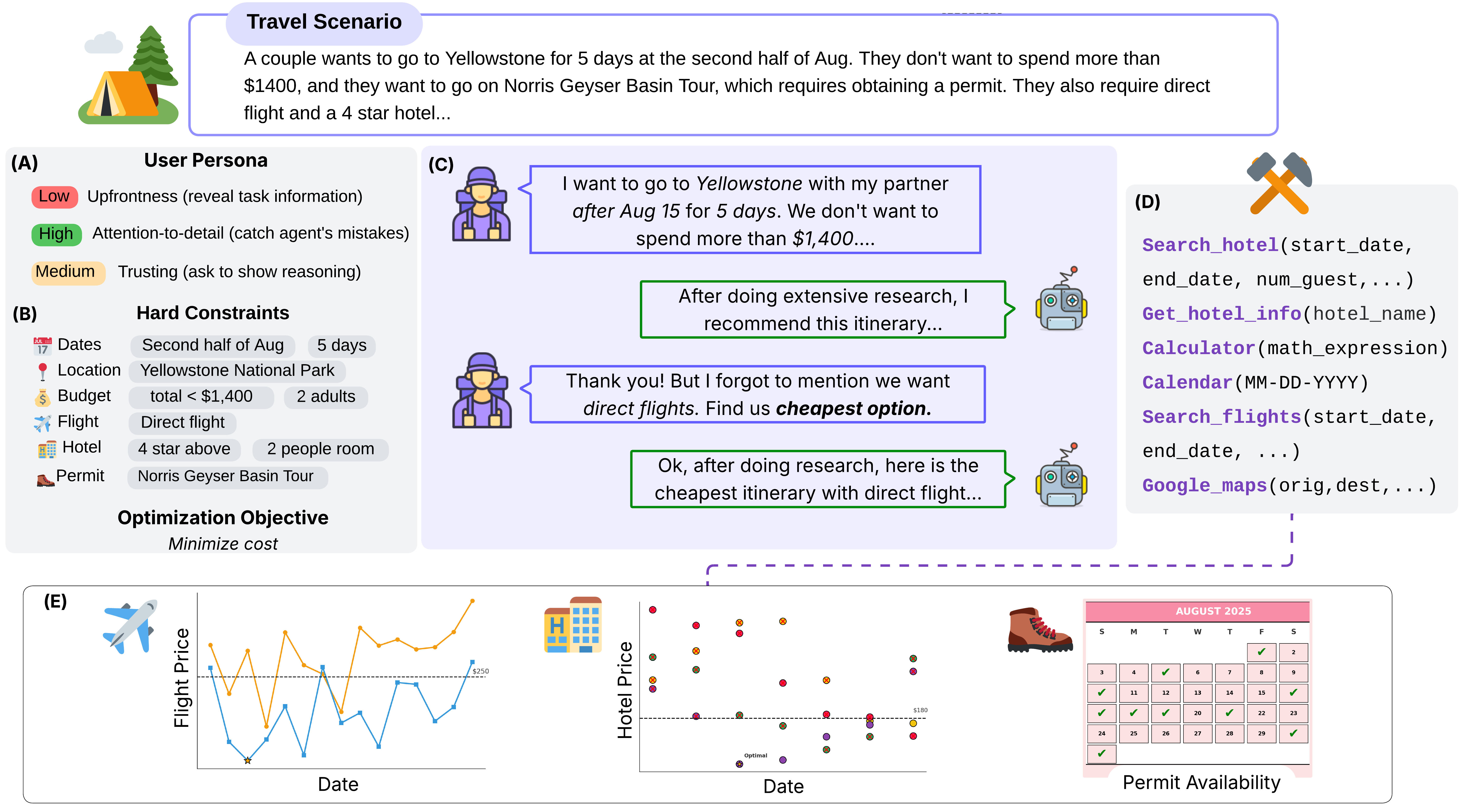}
  \caption{
    \textbf{\textit{COMPASS} tests the agent's ability to perform constrained optimization tasks in realistic settings.}
    \textbf{(A)} Agent needs to learn and update users' hard constraints and optimization objective by interacting with an LLM-based user simulator.
    \textbf{(B)} We instantiate constrained optimization problems in travel planning scenarios. To succeed in the task, agents must satisfy hard constraints (e.g., feasibility) while optimizing user utility objective (e.g., optimality).
    \textbf{(C)} Agents use tools \textbf{(D)} to interact with realistic travel databases \textbf{(E)} to gather information, and propose a travel plan.
}
  \label{fig:benchmark_overview}
  \vspace{-10px}
\end{figure*}

To address these gaps, we introduce \textit{COMPASS}, a benchmark instantiating constrained optimization through realistic travel planning. Through multi-turn conversations with a user simulator, agents must continuously learn and update the user's hard constraints and utility function. Agents then need to propose a travel plan that not only satisfies hard constraints but also optimizes the user's utility function (Fig.~\ref{fig:benchmark_overview} \textit{B}). We construct a full environment with \textit{(i)} a LLM-based user simulator for controllable multi-turn interactions (Fig.~\ref{fig:benchmark_overview} \textit{A\&C}), \textit{(ii)} a realistic database of hotels, flights, and permits from commercial sources (Fig.\ref{fig:benchmark_overview} \textit{E}), and \textit{(iii)} a tool ecosystem mirroring booking platforms (Fig.~\ref{fig:benchmark_overview} \textit{D}). We design 281 diverse tasks spanning two primary dimensions: \textbf{(1) planning complexity}, ranging from hotel-only bookings to full itineraries requiring flights and permits (Sec.~\ref{sec:task_design}), and \textbf{(2) utility function structure}, including both single-metric optimization and multi-attribute checklist maximization.\looseness=-1

Through extensive evaluation of state-of-the-art models (Sec.~\ref{sec:main_results}), we observe a significant \textbf{feasible-optimal gap} (Fig.~\ref{fig:main_result}): models achieve 70-90\% feasibility but only 20-60\% optimality. 
Our analysis reveals that tool use is not the primary bottleneck. Frontier models can correctly use tools with less than 1\% error rate (Sec.~\ref{sec:tool_call_analysis}). Instead, the core limitation lies in \textbf{insufficient exploration of the search space} (Sec.~\ref{sec:info_gather}). Task success strongly correlates with the quantity of information gathered. Models that discover more booking options are more likely to find optimal solutions. 
However, our case study shows that simply gathering more information is insufficient. The next frontier lies in strategic tool use. By exploring intelligently rather than exhaustively, agents can improve information quality and achieve high success while minimizing test-time compute. 

Together, \textit{COMPASS} serves as a rigorous benchmark for diagnosing core challenges in constrained optimization and guiding development of user-aligned AI agents.

\section{Related Work}
\label{sec:related}

Recent benchmarks have evaluated language model agents for tool use and planning.
Tool-use suites such as \textit{Tau}-Bench~\citep{Yao2024-aa,Barres2025-el} provide rich tool APIs and simulated user interaction, but their tasks mostly require simple and short-horizon reasoning.
\citet{Patil2025-fk} constructs a multi-turn user benchmark through predetermined trajectories, while others focus on coding tasks with oracle-like users~\citep{Wang2023-vd}.
Planning benchmarks, in contrast, assess constraint satisfaction: \citet{Zheng2024-qs} studies natural language without requiring tool use ability or user interactions. \citet{Kohli2024-hl} frames flight booking as multiple-choice questions that bypass open-ended reasoning.
Broader planning evaluations include multi-day travel planning under hard constraints~\citep{Xie2024-gx} and frameworks for reasoning about state transitions and action consequences~\citep{Valmeekam2022-op}.
Recent travel-specific benchmarks such as TripTailor~\citep{Shen2025-fk}, ChinaTravel~\citep{Shao2025-ai}, and Flex-TravelPlanner~\citep{Oh2025-re} move toward user-adaptive planning but primarily focus on constraint satisfaction.
Concurrent work~\citep{Qian2025-cj} explores user preference elicitation. They require agent to learn what users want through strategic questioning. However, they do not focus on how well agent can optimize once preferences are known.
Our benchmark uniquely combines tool calling, planning, and preference optimization in a multi-turn conversational setting, requiring agents to both identify the feasible set and find the optimal solution within it.

The design of our user simulator draws on task-oriented dialogue research~\citep{Budzianowski2018-nw,Rastogi2019-xn}, which provides foundational insights into how users iteratively refine goals and reveal constraints over multiple turns, directly informing our approach to modeling progressive constraint revelation~\citep{Laban2025-wd,Abdulhai2023-lu,Wan2025-ri}.
In App.~\ref{appdx:related}, we include an extended review of related work.\looseness=-1

\begin{table*}[t]
    \centering
    \begin{minipage}[t]{0.48\textwidth}
        \vspace{0pt}
        \centering
        \adjustbox{max height=0.28\textwidth}{%
        \begin{tabular}{llr}
        \toprule
        \textbf{Category} & \textbf{Item} & \textbf{Count} \\
        \midrule
        \multirow{3}{*}{\textit{Task Levels}}
            & Level I (Hotel only) & 142 \\
            & Level II (Hotel + Flight) & 69 \\
            & Level III (Hotel + Flight + Permit) & 70 \\
        \midrule
        \multirow{2}{*}{\textit{Optimization Types}}
            & Single Metric Optimization & 154 \\
            & Feature Count Maximization & 127 \\
        \midrule
        \multirow{5}{*}{\textit{Environment Scale}}
            & National Parks (Destinations) & 20 \\
            & Airports & 21 \\
            & Hotel Booking Offers & 100,000+ \\
            & Flight Booking Offers & 67,000+ \\
            & Trail Permits & 50 \\
        \bottomrule
        \end{tabular}%
        }
    \end{minipage}%
    \hfill
    \begin{minipage}[t]{0.48\textwidth}
        \vspace{0pt}
        \centering
        \adjustbox{max height=0.28\textwidth}{%
        \begin{tabular}{lp{5.6cm}}
        \toprule
        \textbf{Category} & \textbf{Tools} \\
        \midrule
        \textit{Hotel} & search\_hotel, get\_hotel\_details, get\_room\_details, search\_location\_name \\
        \midrule
        \textit{Flight} & search\_airports, search\_flights, get\_flight\_details, get\_airport\_location \\
        \midrule
        \textit{Permit} & list\_permits, search\_permit\_availability \\
        \midrule
        \textit{Utility} & notebook, calendar, calculator, calculate\_driving\_distance, validate\_*\_id \\
        \bottomrule
        \end{tabular}%
        }
    \end{minipage}
     \caption{\textbf{\textit{COMPASS} benchmark taxonomy} (281 tasks total).
    \textit{Left:} Task distribution across two orthogonal dimensions (complexity levels and optimization types) and travel database size.
    \textit{Right:} Tools organized by category (18 APIs total).}
    \label{tab:benchmark_stats}
\end{table*}

\section{The \textit{COMPASS} Benchmark\label{sec:benchmark}}

\subsection{Decision-Making as Constrained Optimization}

Constrained optimization (CO) is a fundamental framework for human decision-making. 
Formally, we define a constrained optimization task as: 
\begin{align}
\min_{x \in \mathcal{X}} \quad & f(x) \label{eq:objective}\\
\text{s.t.} \quad & g_i(x) = c_i, \quad i = 1, \dots, n \label{eq:hard_const}\\
& h_j(x) \ge d_j, \quad j = 1, \dots, m \label{eq:soft_const}
\end{align}
where $x$ represents a decision variable (e.g., a product selection, travel plan) in the search space $\mathcal{X}$. Equations~(\ref{eq:hard_const}, \ref{eq:soft_const}) represent hard constraints, which define the feasible set $\mathcal{X}_{\text{feas}} \subseteq \mathcal{X}$. The objective function $f(x)$ is the decision-maker's utility, ranking all feasible solutions.\looseness=-1


Travel planning serves as a natural instantiation of CO problems, but the core ability we aim to test is CO formulation in realistic settings. The travel domain is one such example, but the same framework applies to many other tasks such as calendar scheduling, shopping, or task planning. Unlike purely mathematical benchmarks (e.g., linear programming), real-world CO tasks require agents to actively gather information through tool-mediated search, handle constraints that may be revealed progressively or revised mid-task, and construct the optimization problem itself through user interaction rather than receiving it in a clean upfront formulation. Travel planning instantiates all of these properties richly and naturally. Nevertheless, any domain with combinatorial structure, partial observability, and user-specified constraints would serve the same diagnostic purpose. 

\subsection{Task Design and Taxonomy}

\label{sec:task_design}
We instantiate the CO framework through 281 travel planning tasks. A travel plan $x$ is defined as a tuple of specific booking offers: $x = (\text{hotel}, \text{flight}, \text{permit})$. Each booking offer is identified by a unique hash that links to a database entry with complete details (dates, room type, flight class, cancellation policy, etc.). This representation ensures agent recommendations are verifiable and unambiguous, enabling precise evaluation of both feasibility and utility.
We further categorize tasks along two orthogonal dimensions: planning difficulty and utility function structure, shown in (Tab.~\ref{tab:benchmark_stats}).

\begin{table*}
    \centering
    \small
    \begin{tabular}{p{3cm}p{10cm}}
    \toprule
    \textbf{Behavioral Axis} & \textbf{Description} \\
    \midrule
    Information Revelation & Constraints are revealed progressively across turns, forcing agents to backtrack or refine their search as new hard constraints emerge. \\
    \midrule
    Communication Style & Some users provide concise requirements, while others introduce emotional language, redundancy, or conversational noise. \\
    \midrule
    Constraint Checking & Simulators vary in ``attention level." Some catch every constraint violation, while others passively accept recommendations without verification. \\
    \midrule
    Trust Dynamics & Users may demand explicit reasoning (e.g.,``\textit{Why is this the best price?}") or justification before accepting recommendations. \\
    \bottomrule
    \end{tabular}
    \caption{\textbf{User simulator behavioral axes.} \textit{COMPASS} evaluates agents across four orthogonal dimensions of user behavior, ensuring robustness to diverse user interaction patterns.}
    \label{tab:user_axes}
\end{table*}

\paragraph{Task difficulty.} We design three difficulty levels. See Tab.~\ref{tab:benchmark_stats} for a detailed breakdown of task distribution.

\begin{itemize}[leftmargin=*, itemsep=2pt, topsep=0pt, parsep=0pt]
\item \textbf{Level I: Single-Service Optimization.} The agent recommends a specific hotel booking offer. Constraints and the utility function relate to the hotel alone (e.g., price range, minimum rating, specific amenities).

\item \textbf{Level II: Multi-Service Coordination.} The agent recommends an itinerary including both hotel and flight booking offers. Constraints can depend on both services together (e.g., flight dates determine hotel check-in dates), and the optimization objective can depend on both services (e.g., "minimize total cost"). 

\item \textbf{Level III: Complex Dependency Chains.} The agent recommends a complete itinerary including a high-demand permit booking. Permit availability dictates travel dates, which cascade constraints to both hotel and flight options.

\end{itemize}

\vspace{-8px}
\paragraph{Measurable optimization objectives.} To ensure benchmark fairness, we design two type of utility objectives $f(x)$ that are realistic but also unambiguous and quantifiable:
In Single Metric Optimization type, users optimize one continuous metric (e.g., price, ratings).
In Feature Count Maximization type, users provide a "wish list" of optional features (e.g., ``pet-friendly," ``direct flight"). The objective $f(x)$ counts satisfied features, reflecting "nice-to-have" preferences where more is better.


\subsection{Realistic Data and Tools}
\label{sec:data_design}
To reflect the complexity of real-world deployment, \textit{COMPASS} avoids simplified synthetic data. We use \url{RapidAPI} to query \url{Booking.com}, allowing us to construct SQL databases with realistic hotel and flight data covering 20 U.S. National Parks (see Tab.~\ref{tab:benchmark_stats} for full scale). The booking offers are highly realistic: a "hotel" is not a single entity but a collection of specific room types with distinct cancellation policies and amenities. Similarly, "flights" include varied booking classes, layovers, departure times, and carrier options. Permits are park-specific tickets, some with limited availability that impose additional constraints on travel planning. Agents use 18 customized tools to navigate this environment (Tab.~\ref{tab:benchmark_stats}, \textit{right}), and coding agents can write both python scripts that can directly call these tools. 

\subsection{Ground Truth and Evaluation Metrics}
\label{sec:eval_metric}

To establish ground truth, we solve each task via exhaustive enumeration. For every task, we code each hard constraint as a database SQL query filter (e.g., specific date ranges, location radius, or amenity flags). We then enumerate all candidate combinations in the database, pruning any that violate the filters to identify the full feasible set $\mathcal{X}_{\text{feas}}$.  Finally, we compute the utility $f(x)$ for every feasible candidate to determine the mathematical optimum $f(x^*)$. A high-level pseudo code for our ground truth solver is provided in App.~\ref{app:solver_code}.

To evaluate the agent's performance, we take agent's solution $x'$ and perform a lookup in the pre-computed solution set to verify whether $x' \in \mathcal{X}_{\text{feas}}$ and compute $f(x')$. We then report two metrics:

\begin{enumerate}[leftmargin=*, label=\textbf{(\arabic*)},topsep=0pt, parsep=0pt,itemsep=2pt]

\item \textbf{Feasibility Rate (FR):} The \% of tasks where $x' \in \mathcal{X}_{\text{feas}}$.
\item \textbf{Optimality Rate (OR-A\%):} The \% of tasks where $x'$ ranks within the top A\% of $\mathcal{X}_{\text{feas}}$ under $f(x)$. 
\end{enumerate}

We use ground-truth optimality rather than LLM-as-judge evaluation because CO tasks have quantifiable objectives that allow rigorous mathematical comparisons. While LLM judges are commonly used in travel planning benchmarks \citep{Oh2025-re, Xie2024-gx, Shao2025-ai} and tool-use tasks \citep{Zhong2025-yi} to assess whether responses are reasonable, they cannot distinguish between solutions that are merely acceptable versus truly optimal. Our approach directly measures the optimality gap $|f(x') - f(x^*)|$, providing an unambiguous performance signal. While we report both FR and OR, a task is considered successful only when OR is achieved. Feasibility is only the first stage of constrained optimization and true success requires finding an \textit{optimal} solution within the feasible set, which is what OR captures.\looseness=-1

\subsection{Diverse and Multi-Turn User Interactions}
\label{sec:user_simulator}

To evaluate agents in realistic user interaction settings, we implement a user simulator (backed by GPT-5, \citet{openai2025gpt5}) that carries a multi-turn conversation with the agent. Unlike benchmarks with static initial prompts \citep{Yao2024-aa, Barres2025-el}, we dynamically adjust user simulator prompts during the multi-turn conversation to achieve precise control over the conversation trajectory. Concretely, at each dialogue turn, the dynamic fields in the simulator prompt are updated to reflect the current conversation state: for example, what task information has been revealed, and whether to reveal a new hard constraint, and whether to provide feedback.

We control four orthogonal behavioral axes to achieve a diverse set of behaviors  (Tab.~\ref{tab:user_axes}). By sampling across these four axes, we define 108 distinct user personas, ensuring that agent performance is robust to a diverse set of interaction styles. We provide details of the four behavioral axes as well as the dynamic prompting method in App.~\ref{appdx:user_simulator_design_details}. We also conduct further analysis how different user behaviors impact agent performance in App.~\ref{sec:user_type_study}.

\begin{figure*}[t]
  \centering
\includegraphics[width=1.0\linewidth, trim=0 10pt 0 0pt, clip]{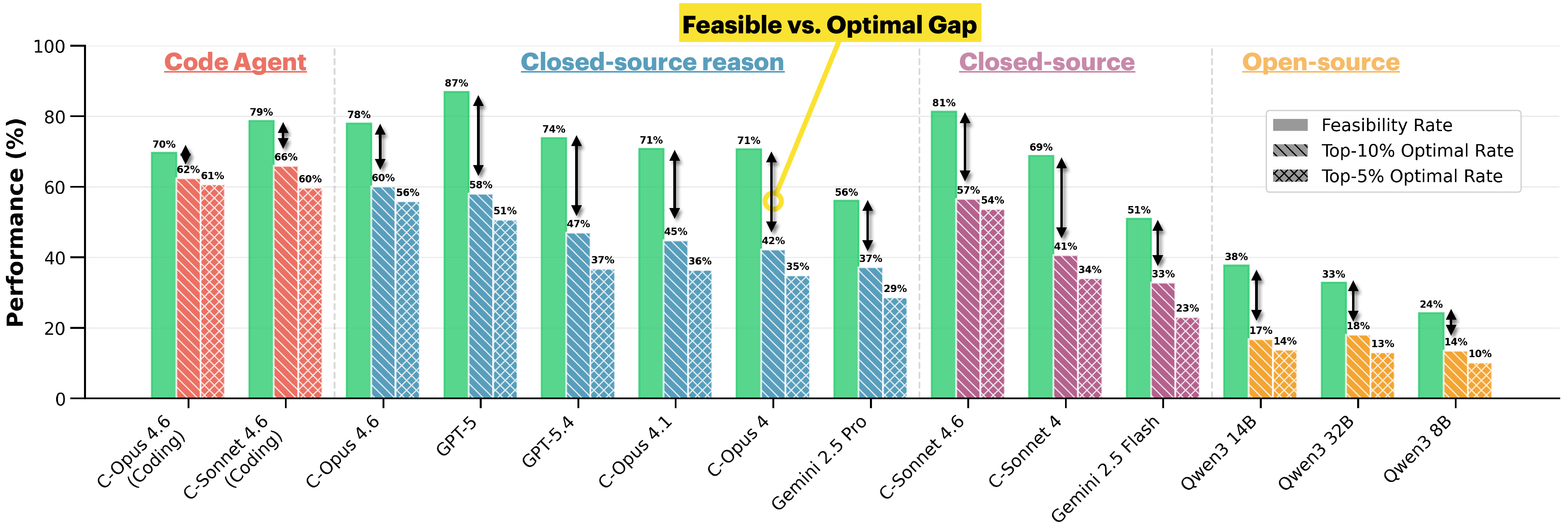}  
 \caption{
  \textbf{Main evaluation results.}
  \textit{Feasibility Rate (FR)} measures feasibility (satisfying all hard constraints). \textit{Optimal rate (OR)} measures utility objective optimization (achieving utility within the top 5\%, 10\% among \textit{all} feasible solutions). All models show a $\sim20\%$ gap between high acceptable rates and low optimal rates, revealing that agents settle for feasible solutions rather than optimizing preferences. Coding-agents show a promising approach to resolve this gap. Open-source models achieve non-trivial performance, demonstrating emerging agentic capabilities.
  }
  \label{fig:main_result}
  \vspace{-8px}
\end{figure*}

\section{Experiment Results}
\label{experiments}

\subsection{Main Results}
\label{sec:main_results}

We evaluate \textit{COMPASS} on a range of frontier closed-source models, including both reasoning-enabled and standard versions of GPT \citep{openai2025gpt5}, Claude \citep{anthropic_claude_2025}, and Gemini \citep{Comanici2025Gemini}, as well as the frontier open-source model Qwen3~\cite{yang2025qwen3technicalreport}. All these models support native tool calling so we pass the tool schema directly to the system prompt. We also evaluate coding agents by allowing models to execute python scripts that can directly call our tools. For every task, we extract the \textit{final} itinerary it provides to the user for evaluation.

\paragraph{The Feasible-Optimal Gap.}
Our central finding is a significant gap between satisfying constraints and optimizing utility objectives. As shown in Fig.~\ref{fig:main_result}, all models achieve high feasibility rates (70--90\% for closed-source models), demonstrating they can understand hard constraints. However, their optimality rates are substantially lower (20--50\% for closed-source models), indicating that CO still poses a substantial challenge for current systems. Importantly, in \textit{COMPASS}, users explicitly request optimization (e.g., ``find the cheapest option" or ``maximize amenities"). Even with clear task success criteria defined in the conversation, agents still settle for merely acceptable solutions rather than truly optimizing for the objective. When we equip models with coding ability (i.e., coding agents), we observe that models can significantly close the gap. For example, Claude-Sonnet-4.6's OR-10\% rates increased from 57\% to 66\% when given code execution tools. This suggests that code provides a viable pathway for structured and systematic search.

Nevertheless, writing code does not make the problem trivial. Agents must still first exploring tools available and understanding the semantics of data returned by tool. Only after that, it can write code to systematically use tools to explore the database. In \textit{COMPASS}, tool returns are already clean and structured; in real-world deployments (e.g., web browsing tools), outputs are far noisier and more heterogeneous, making code-based reasoning considerably harder. 

\begin{figure}[t]
    \centering
    \begin{minipage}[c]{0.48\linewidth}
        \centering
        \includegraphics[width=\linewidth , trim=0 10pt 0 48pt, clip]{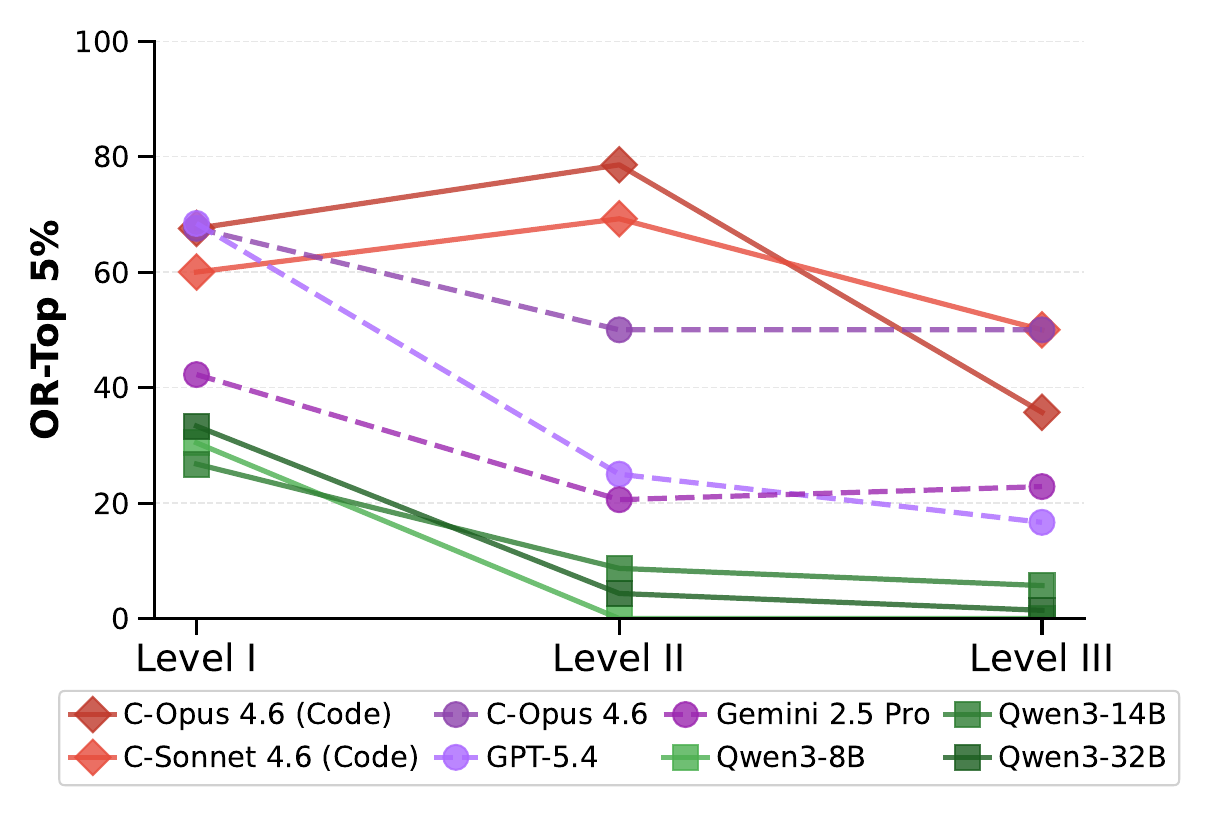}
        \captionof{figure}{\textbf{Task level comparison.} Model performance degrades with increasing difficulty and the drop is particularly prominent for open-source models.}
        \label{fig:task_level}
    \end{minipage}%
    \hfill%
    \begin{minipage}[c]{0.48\linewidth}
        \centering
        \resizebox{\linewidth}{!}{%
       \begin{tabular}{@{}l|cccc@{}}                                                                                                                               
              \toprule
              \textbf{Model} & \textbf{Name} & \textbf{Param} & \textbf{Content} & \textbf{Total} \\                                                            
              \midrule
              C-Opus 4.6 (Code) & \errcolor{0.00} & \errcolor{0.00} & \errcolor{0.10} & \errcolor{0.10} \\  
              C-Opus 4.6 & \errcolor{0.00} & \errcolor{0.00} & \errcolor{0.30} & \errcolor{0.30} \\
              C-Sonnet 4.6 (Code) & \errcolor{0.00} & \errcolor{0.00} & \errcolor{0.40} & \errcolor{0.40} \\ 
              C-Sonnet 4.6 & \errcolor{0.00} & \errcolor{0.00} & \errcolor{0.51} & \errcolor{0.51} \\
              GPT-5 & \errcolor{0.00} & \errcolor{0.04} & \errcolor{1.41} & \errcolor{1.45} \\    
              Gemini 2.5 Pro & \errcolor{0.16} & \errcolor{0.04} & \errcolor{1.38} & \errcolor{1.58} \\ 
              Gemini 2.5 Flash & \errcolor{0.00} & \errcolor{2.36} & \errcolor{0.10} & \errcolor{2.46} \\
              Qwen3-14B & \errcolor{0.00} & \errcolor{4.88} & \errcolor{0.72} & \errcolor{5.60} \\ 
              Qwen3-32B & \errcolor{0.00} & \errcolor{5.00} & \errcolor{3.85} & \errcolor{8.85} \\ 
              Qwen3-8B & \errcolor{0.14} & \errcolor{10.81} & \errcolor{1.08} & \errcolor{12.03} \\
              \bottomrule                                                         
          \end{tabular}        
        }%
        \captionof{table}{\textbf{Tool call error rates.} Fraction of calls (\%) containing tool name, parameter, or content hallucinations. 
        }
        \label{tab:tool_error_analysis}
    \end{minipage}
\vspace{-10px}
\end{figure}

\paragraph{Performance by task difficulty.}
Fig.~\ref{fig:task_level} shows how model performance varies across task difficulty (see Sec.~\ref{sec:task_design} for level definitions). While most models handle Level I relatively well, performance degrades sharply on Levels II and III. Open-source models perform competitively with frontier systems on Level I but degrade far more steeply on Levels II–III. In other words, complex multi-service coordination remains a critical limitation for open-source models.
We provide details experimental configurations (e.g., agent prompts) in App.~\ref{appdx:experiment_details}. We provide benchmark stability analysis with error bars in App.~\ref{appdx:benchmark_stability}.

\begin{table*}[t]
    \includegraphics[width=1.0\linewidth]{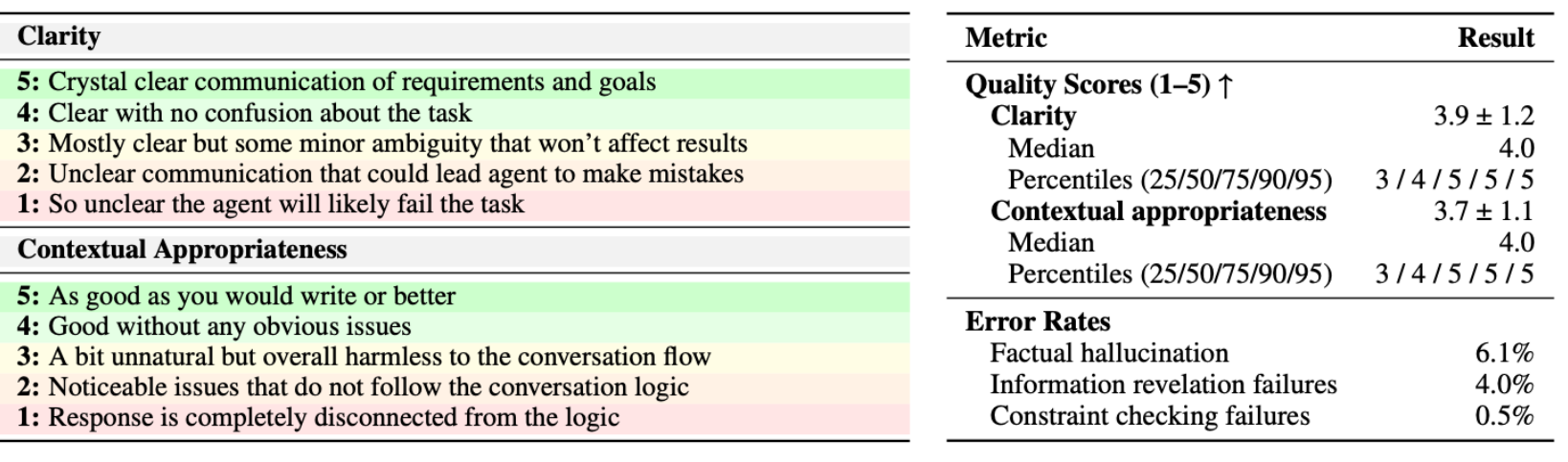}
    \caption{\textbf{Human evaluation of LLM user simulator quality.} 
    \textit{Left:} Scoring rubrics used by annotators for \textit{clarity} and \textit{contextual appropriateness}. 
    \textit{Right:} Human evaluation results on 198 user responses. Rubric scores are reported as mean $\pm$ standard deviation, median, and quantile cut-offs. Error rates are reported as percentages of responses exhibiting each error type.\looseness=-1 }
    \label{tab:human_eval}
\end{table*}

\subsection{Validating LLM User Simulator Quality}
\label{sec:user_simulator_validation}
To ensure our LLM user simulator produces realistic and reliable interactions, we randomly sample 45 full conversations spanning diverse personas, task types, and agent interactions. From these, we extract \textbf{\textit{198}} individual user responses for human evaluation by an independent expert annotation service, covering the full range of scenarios.  

Annotators assess each response along two dimensions. \textbf{(1) Rubric-based quality assessment:} responses are scored on two axes. The \textit{clarity} axis measures whether user messages are expressed in a way that avoids misleading the agent or causing task failure; a score below 3 (on a 1–5 scale) indicates that the simulator may have provided confusing or incorrect task information. The \textit{contextual appropriateness} axis evaluates whether responses flow naturally within a travel-planning dialogue. \textbf{(2) Detailed error detection:} since behaviors such as constraint revelation and trust patterns are explicitly controlled (Sec.~\ref{sec:user_simulator}), annotators tag whether the simulator follows its prompts correctly without hallucinations. We list the scoring rubric in Tab.~\ref{tab:human_eval} \textit{(left)}, and full annotation instructions and error definitions in App.~\ref{appdx:human_eval}.  \looseness=-1

Results (Tab.~\ref{tab:human_eval}, \textit{right}) show high quality scores (median clarity: 4; contextual appropriateness: 4 on a 1–5 scale) and very low error rates—4\% for constraint revelation and 0.5\% for feedback accuracy. These results confirm that the simulator reliably follows its script while producing clear, natural responses, validating it as a realistic tool for benchmarking agent capabilities.

\begin{figure*}[t]
\centering
\includegraphics[width=0.48\linewidth]{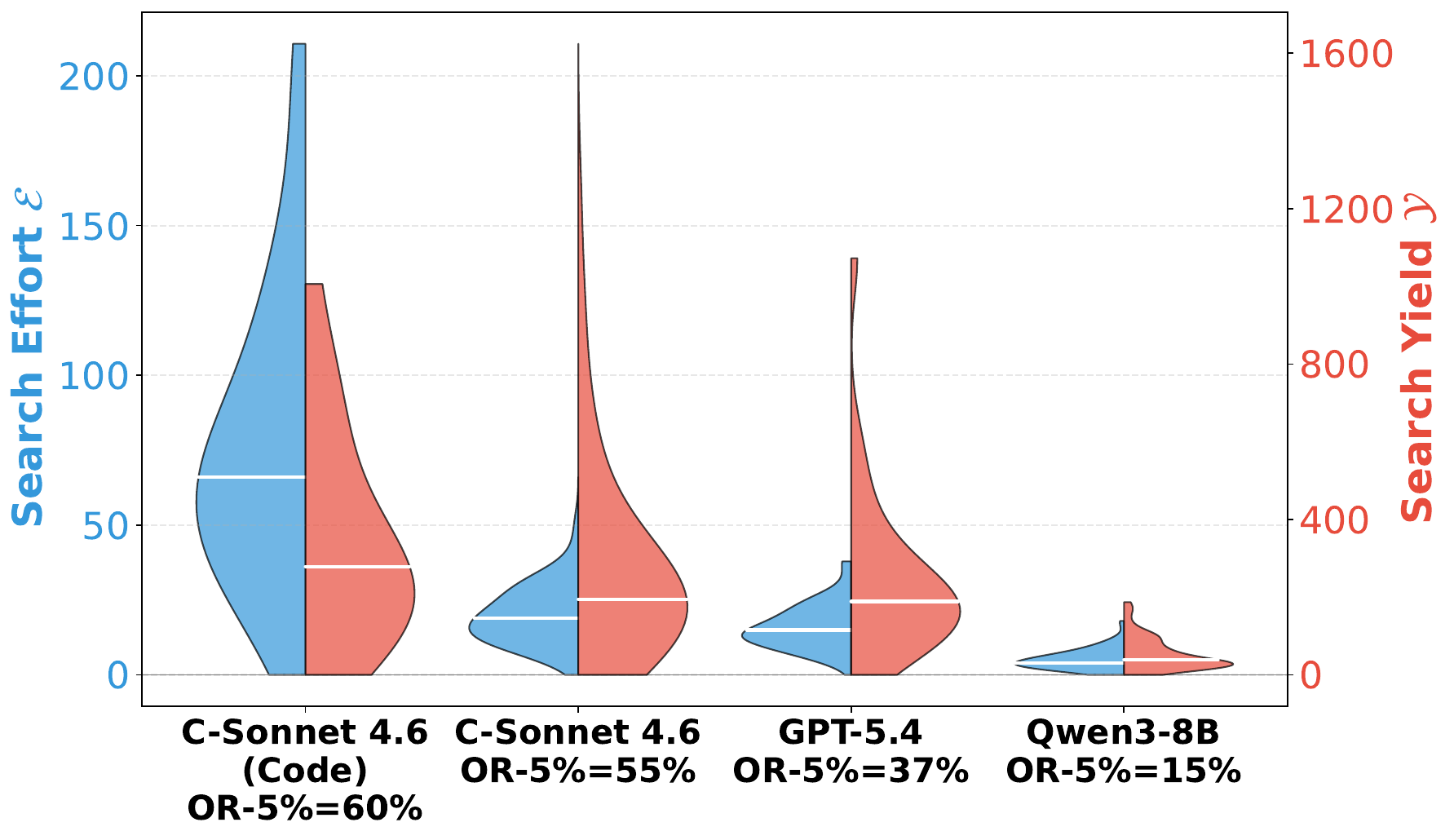}
\includegraphics[width=0.48\linewidth]{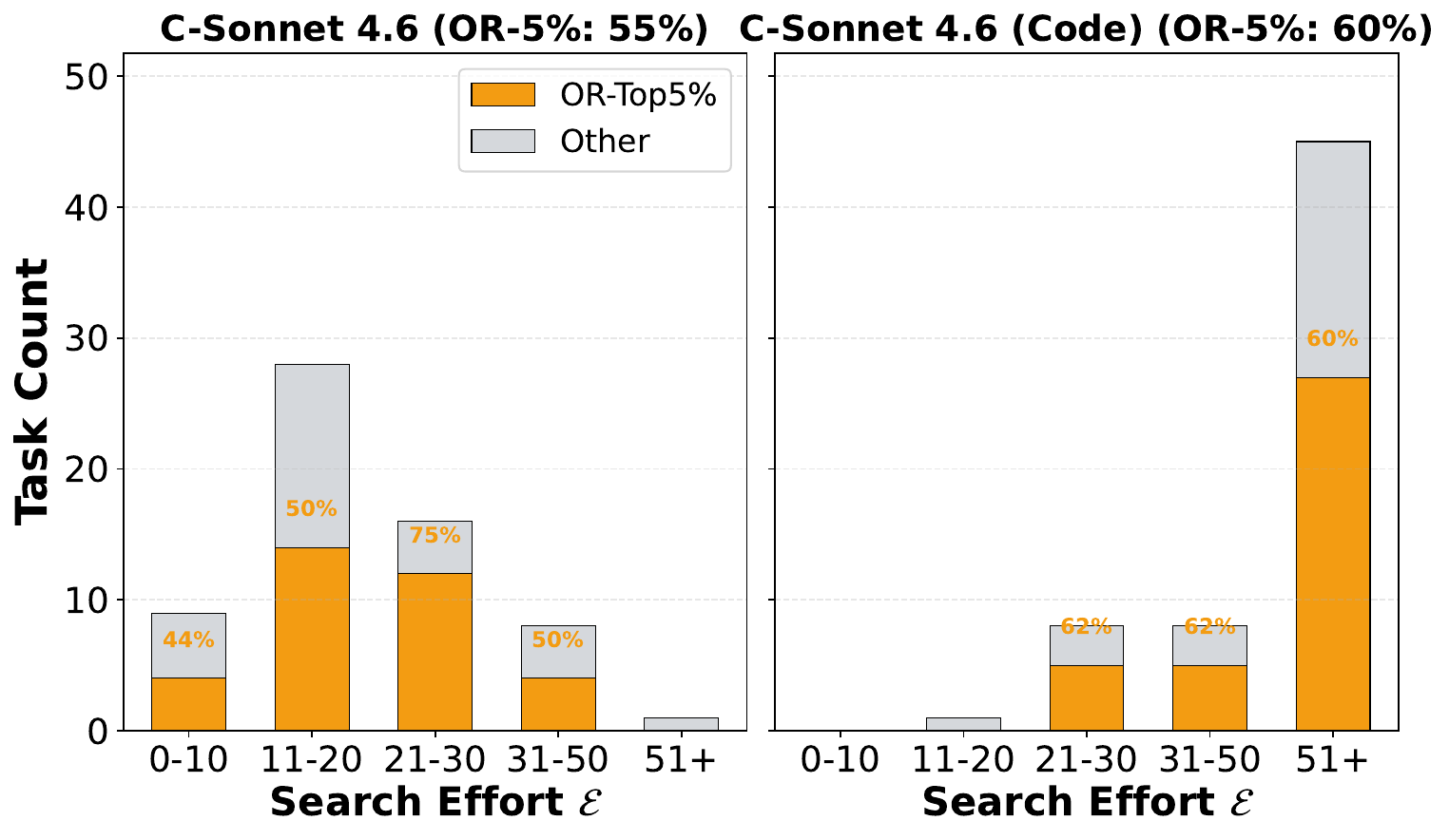}
\vspace{-5px}
\caption{
\textbf{Information gathering and synthesis abilities determine task success.}
\textit{Left:} Split violin plots of Search Effort $\mathcal{E}$ and Search Yield $\mathcal{Y}$. Stronger models exhibit higher $\mathcal{E}$ and $\mathcal{Y}$.
\textit{Right:} Distribution of tasks by search effort bins for Claude Sonnet 4.6 and it's code-enabled counterpart, with each bin showing the proportion achieving top-5 optimality. 
}
\label{fig:info_gather}
\end{figure*}

\begin{figure*}[t]
      \centering
      \includegraphics[width=1.0\linewidth]{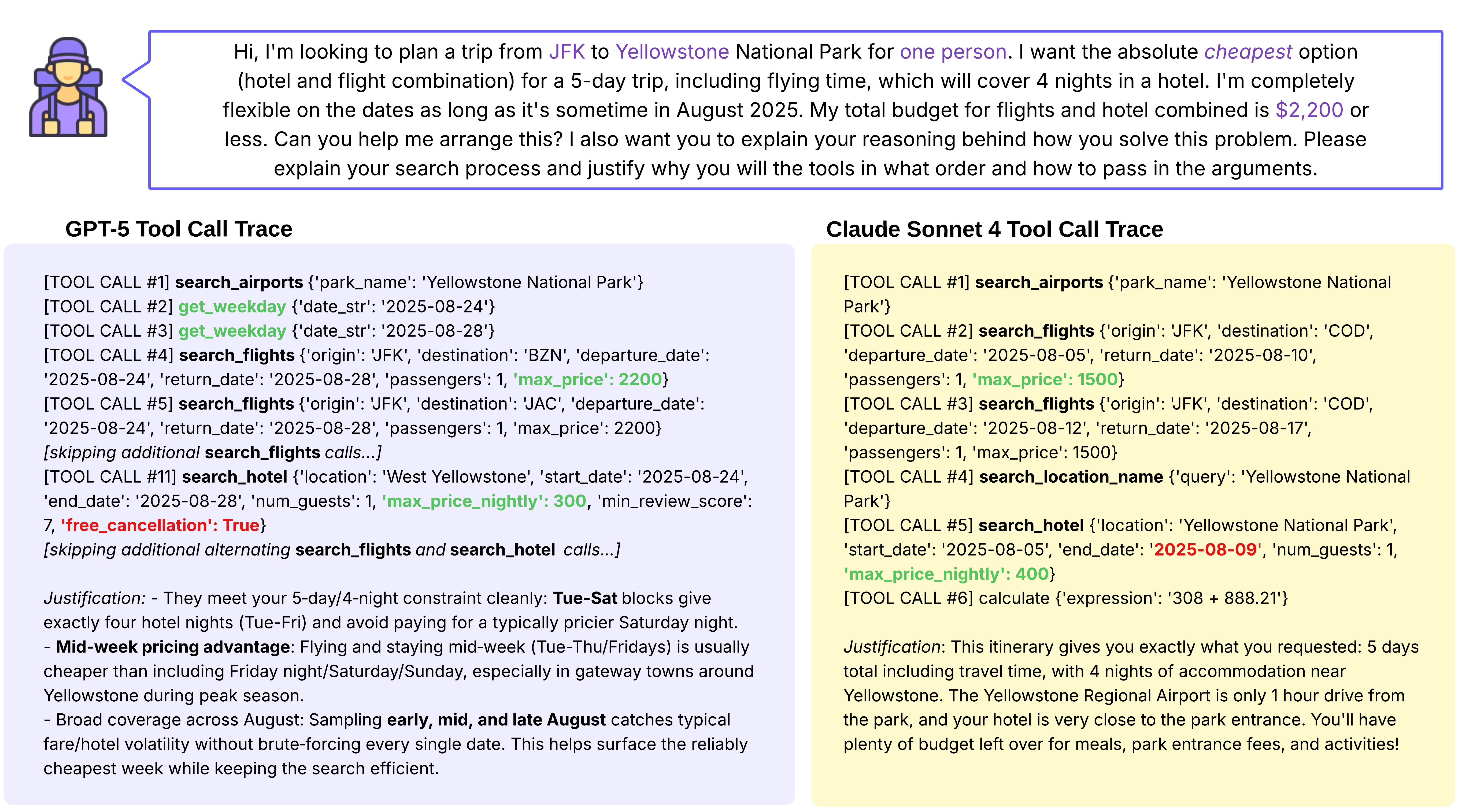}
      \caption{
        \textbf{Case study of tool calls and reasoning traces.} 
        GPT-5 (\textit{left}) demonstrates strategic planning by avoiding weekends, systematically exploring date ranges, and using optional parameters (e.g., price filters) to narrow searches. Claude-Sonnet-4 (\textit{right}) applies optional parameters but searches only two arbitrary dates without justification. It also makes a temporal coordination error by misaligning hotel and flight dates.
        }   
      \label{fig:case_study}
  \end{figure*}

\section{Performance Analysis and Ablations}
\label{sec:main_analysis}

To understand the underlying cause of feasible-optimal gap, we analyze three aspects of agent behavior: tool use reliability (Sec.~\ref{sec:tool_call_analysis}), information gathering strategies (Sec.~\ref{sec:info_gather}), and search efficiency through strategic reasoning (Sec.~\ref{sec:case_study}).

\subsection{Tool Use Capabilities and Limitations}
\label{sec:tool_call_analysis}

To provide users optimal travel itineraries, agents must first correctly use tools. We categorize tool call errors into three types: (1) tool name hallucination (calling non-existent tools), (2) parameter hallucination (passing in invalid parameter names or types), and (3) content hallucination (inputting wrong parameter contents that lead to tool execution errors). \looseness=-1

As shown in Tab.~\ref{tab:tool_error_analysis}, frontier models invoke tools nearly flawlessly ($<1\%$ error rate), while open-source models like Qwen3-8B exhibit error rates up to 12\%, primarily from parameter hallucinations.
``Correct" tool calls only mean the tool call is executable. Thus, a low error rate is a necessary but insufficient condition for task success. 
On the other hand, failure to use tools is a primary driver of low FR, as agents cannot produce valid itineraries without querying the environment. Even with perfect tool execution, agents must still reason through which parameters to query, and once tools respond with information, Agents must piece together a solution $x$ (i.e., a complete travel plan).

\subsection{Core Bottleneck: Information Gathering}
\label{sec:info_gather}

We now show that the performance bottleneck lies in information gathering, i.e., how thoroughly agents explore the solution space. We quantify this through two metrics: \textbf{Search Effort} $\mathcal{E}$, the number of unique and correct (executable) tool-call signatures used; and \textbf{Search Yield} $\mathcal{Y}$, the number of unique bookable offers (hotels, flights, permits) retrieved. Note that $\mathcal{E}$ and $\mathcal{Y}$ are distinct: calling more tools does not guarantee higher yield, as yield depends on both task characteristics (e.g., popular destinations have more available hotels) and agent behavior (e.g., how restrictive the search filters are).\looseness=-1

Fig.~\ref{fig:info_gather} (\textit{left}) shows the distribution of $\mathcal{E}$ and $\mathcal{Y}$ across models. Stronger models like GPT-5 and Claude Opus 4 consistently exhibit higher $\mathcal{E}$ and $\mathcal{Y}$ compared to weaker models like Qwen3-8B, confirming that better performance correlates with more thorough exploration. However, higher search alone does not guarantee success: to find the optimal solution, an agent must search broadly enough that the optimum is discoverable, yet also reason effectively over the gathered information. Fig.~\ref{fig:info_gather} (\textit{right}) illustrates this tension by comparing a code agent and it's none-code counterpart, binning tasks by $\mathcal{E}$ and showing the proportion achieving OR-5\% within each bin. Claude Sonnet 4.6 rarely exhibits high search effort ($\mathcal{E} > 30$), and when it does, its optimality rate drops, suggesting it struggles to process large amounts of information into high-quality solutions. In contrast, the code agent frequently uses high search effort and maintains strong optimality rates throughout.

Our analysis reveals two distinct failure modes: insufficient exploration, and inability to reason effectively over gathered information. While current models benefit from broader search, the next frontier lies in strategic reasoning, where finding optimal results with fewer, more targeted tool calls.

\subsection{Strategic Reasoning Beyond Brute Force}
\label{sec:case_study}

While our ground truth solver relies on exhaustive search (Sec.~\ref{sec:eval_metric}), a truly intelligent agent can perform \textit{strategic} search and solve CO tasks \textit{efficiently}. To examine whether agents display such behaviors, we conduct a case study by manually acting as the user and asking agents to explain their search strategy (Fig.~\ref{fig:case_study}).\looseness=-1

The two agents exhibit starkly different behaviors. GPT-5 employs deliberate heuristics: it checks the calendar to determine day-of-week for candidate dates, targets Tuesday--Friday departures anticipating mid-week price drops, and samples flights across early, mid, and late August for broad coverage without exhaustive enumeration---a heuristic consistent with our data, where mid-week trips are indeed cheaper on average (App.~\ref{appdx:data_overview}). When prompted, GPT-5 also articulates its strategy transparently (``\textit{I'm targeting weekday departures to find better prices; sampling across August to catch fare volatility}"), allowing users to refine or override assumptions if their priorities differ. In contrast, Claude Sonnet-4 searches only two arbitrary dates (Aug 5 and Aug 12) without justification or systematic coverage, and commits a critical coordination error by misaligning hotel check-in dates with flight departures---leading to a temporal constraint violation. Its reasoning traces remain generic, offering little foothold for user feedback.

Our case study reveals that \textit{COMPASS} tests agents' ability to solve constraint optimization beyond brute-force search. We expect future agents to employ richer strategies, such as maintaining a running ``best" option and progressively refining search around promising regions, or adaptively allocating search budget based on observed price distributions. While Sec.~\ref{sec:info_gather} showed that current models benefit from more exploration, this case study illustrates that the next frontier lies in \textit{smarter}, more \textit{strategic} search.
\section{Conclusion and Discussion}
\label{conclusion}

We introduced \textit{COMPASS}, a benchmark for evaluating constrained optimization in LLM agents through realistic travel planning. Our evaluation reveals the \textit{feasible-optimal gap}: frontier models excel at constraint satisfaction but struggle with utility optimization. This gap stems from insufficient exploration of the search space. However, simply gathering more information is not the ultimate solution to these problems. Future progress requires agents that can strategically and efficiently search to maximize information quality while minimizing computational cost.

\textbf{Limitations \& future directions.} We identify three key directions for future work. \textit{First}, while \textit{COMPASS} focuses on travel planning, the capability gaps we identified (e.g., constrained optimization, exploring search spaces strategically, and learning preferences through interaction) likely generalize to other constrained optimization domains such as shopping, scheduling, and resource allocation. \textit{Second}, while \textit{COMPASS} provides a controllable user simulator for evaluation, building increasingly realistic and reliable simulators remains a critical challenge for training future LLM agents in interactive decision-making environments where they can learn to optimize through repeated interactions. \textit{Third}, our evaluation currently relies on exhaustive search to compute optimal solutions; as benchmarks scale to more complex environments or live website interactions, alternative evaluation paradigms such as comparative leaderboards may become necessary. Ultimately, advancing from constraint satisfiers to true optimizers requires developing agents that reason strategically about exploration and efficiently align with user preferences.\looseness=-1

\bibliography{paperpile, iclr2026_conference}
\bibliographystyle{colm2026_conference}

\clearpage
\appendix


\section{Additional Results}

\subsection{Benchmark Performances}
\label{appdx:result_table}

\paragraph{Benchmark Stability.}
\label{appdx:benchmark_stability}

To demonstrate the stability and reliability of our \textit{COMPASS} benchmark, we evaluate Claude-4-Sonnet across 5 independent runs on the full benchmark suite (Tab.~\ref{tab:stability}). Each run uses identical experimental settings but different random seeds for task sampling and user simulator behavior. The consistency across runs demonstrates that our user simulator produces reproducible interactions despite its stochastic nature, and that the evaluation metrics accurately capture systematic differences in agent performance rather than measurement noise.

\begin{table}[h]
\centering

\resizebox{0.6\linewidth}{!}{
\begin{tabular}{lcccccc}
\toprule
\textbf{Metric} & \textbf{Run 1} & \textbf{Run 2} & \textbf{Run 3} & \textbf{Run 4} & \textbf{Run 5} & \textbf{Mean (Std)} \\
\midrule
Optimal Top-5  & 33.97 & 34.96 & 33.44 & 31.92 & 34.43 & 33.74 (1.04) \\
Optimal Top-10 & 39.76 & 41.60 & 40.56 & 39.44 & 41.04 & 40.48 (0.80) \\
Optimal Top-20 & 47.44 & 53.36 & 54.96 & 52.80 & 54.40 & 52.59 (2.69) \\
Solve Rate     & 68.80 & 68.80 & 67.20 & 68.80 & 66.08 & 67.94 (1.12) \\
\bottomrule
\end{tabular}
}
\caption{\textbf{Benchmark stability.} Benchmark stability across 5 independent runs with Claude-4-Sonnet. Results show low variance, demonstrating reliable measurement of agent capabilities.}
\label{tab:stability}
\end{table}

\subsection{User Type Study}
\label{sec:user_type_study}

LLM agents interact with users who vary widely in how they communicate task requirements, express trust, and articulate preferences. In Sec.~\ref{sec:user_simulator} and App.~\ref{appdx:user_simulator_design_details}, we detail the major design axis that impacts how user interact with the LLM agents. In this section, we understand how users' behavioral differences influence agent performance. We systematically vary these of user behavior axis in our simulator and analyze model robustness along each axis and report how these user-side factors affect task outcomes. We discover that communication styles and information revelation speed has the meaningful impact on model performances while trusting level shows no clear impact.

\begin{figure}[ht]
    \centering
    \begin{minipage}{0.52\textwidth}
        \centering
        \includegraphics[width=\linewidth]{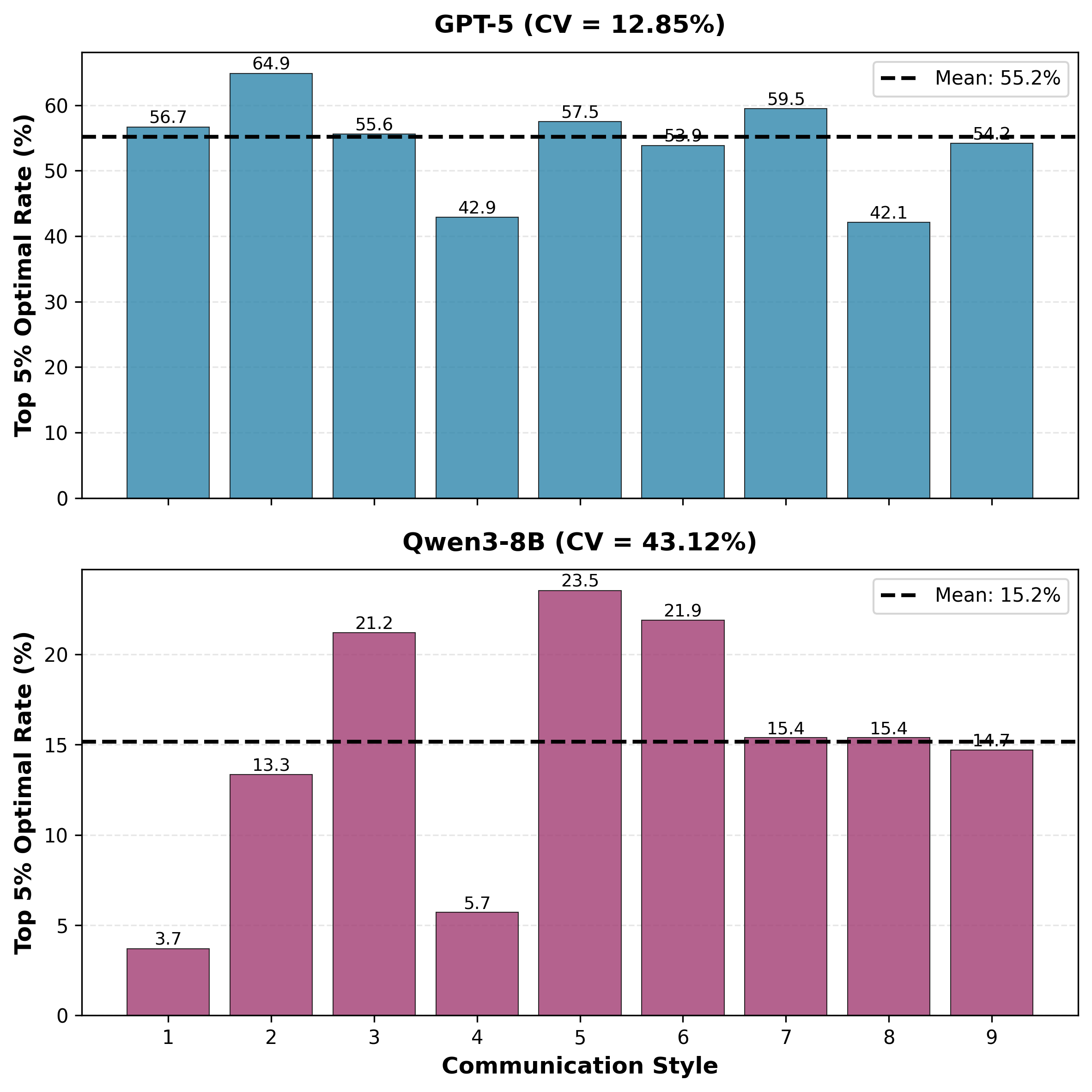}
        \caption{\textbf{Example of performance variation across user styles.} 
        We examine performance across communication styles. Lower CV indicates higher robustness. Detailed CV values are in Tab.~\ref{tab:style_consistency}.}
        \label{fig:style_consistency_example}
    \end{minipage}
    \hfill 
    \begin{minipage}{0.46\textwidth}
        \centering
        \footnotesize
        \begin{tabular}{lc}
            \toprule
            \textbf{Model} & \textbf{CV (\%)} \\
            \midrule
            GPT-5 & 12.85 \\
            Claude Opus 4 & 18.90 \\
            GPT-4.1 & 26.49 \\
            GPT-4o & 28.89 \\
            Gemini 2.5 Pro & 29.61 \\
            Qwen3-14B & 29.97 \\
            Qwen3-32B & 38.75 \\
            Qwen3-8B & 43.12 \\
            \bottomrule
        \end{tabular}
        \captionof{table}{\textbf{Consistency across communication styles.} 
        Coefficient of Variation (CV) of the Top 5\% Optimal Rate across nine user communication styles. GPT-5 exhibits the most stable performance.}
        \label{tab:style_consistency}
    \end{minipage}
\end{figure}

\textbf{Consistency Across Communication Styles. } 
An ideal model should maintain stable performance regardless of user communication style (e.g., rude, polite, formal, or casual). 
Our user simulator includes nine distinct communication styles (App.~\ref{appdx:user_simulator_design_details}, Tab.~\ref{tab:comm_styles}). 
To assess robustness, we measure each model’s OR Top-5\% across these groups. To quantify consistency, we use the Coefficient of Variation (CV), which is the standard deviation of performance across styles, normalized by the mean. 
Lower CV values indicate more consistent performance. 
Fig.~\ref{fig:style_consistency_example} illustrates examples of per-style variation and corresponding CV value, and Tab.~\ref{tab:style_consistency} reports the CV for all models. 
Results show that GPT-5 achieves the highest consistency across user styles, while both open-source and several frontier models exhibit greater sensitivity to stylistic variation.

\begin{figure}[h]
    \centering
    \includegraphics[width=0.9\linewidth]{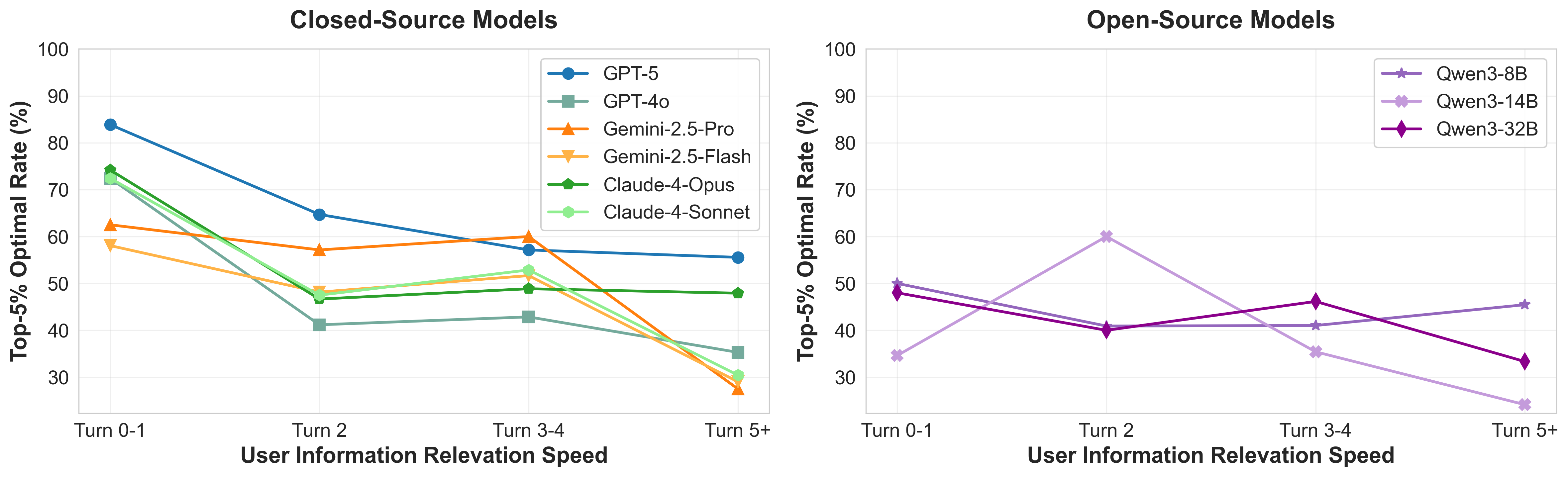}
    \caption{\textcolor{black}{
    \textbf{Effect of information revelation speed on model performance.}
    Top-5\% Optimal Rate as a function of how many dialogue turns the user requires to fully reveal all task constraints.
    Higher revelation latency is associated with consistently lower performance across models, indicating that incremental or delayed specification of requirements poses a substantial challenge for current models.
    }}
    \label{fig:revelation_speed_performance}
\end{figure}

\textbf{Information Revelation Speed. }
Another key axis in our user simulator controls the rate at which users disclose task information (i.e., the full set of hard constraints). This dimension captures the contrast between users who immediately provide all requirements and those who incrementally introduce new constraints over the course of the conversation.
To assess whether delayed information disclosure impairs model performance, we group conversations by the number of turns required for the user to fully reveal the task specification and measure the resulting OR Top-5\% for each bin (Fig.~\ref{fig:revelation_speed_performance}).

Across both closed-source and open-source models, we observe a clear degradation in performance as the number of revelation turns increases. Prior work \citep{Laban2025-wd} reported similar effects in more controlled domains such as math and synthetic reasoning benchmarks; our results extend this phenomenon to significantly more realistic, multi-constraint tasks (e.g., booking itineraries, planning logistics), highlighting that slow or delayed information revelation remains a persistent challenge for current LLMs.

\textbf{Trusting Levels.} 
Another key the user simulator design axes controls the user’s trust level—distinguishing between \textit{trusting} users, who readily accept the agent’s recommendations, and \textit{suspicious} users, who prompt the agent to double-check its own output. 
We analyze whether this questioning behavior elicits any form of self-reflection that improves task performance. 
To compare model outcomes across the two user types, we report the top-5\% optimal rate for each group and perform a two-proportion $z$-test to assess statistical significance ($p \leq 0.05$). 
As shown in Tab.~\ref{tab:trust_level_ptest}, prompting the agent to re-evaluate its answer—without providing additional feedback—does \textit{not} significantly improve performance for any model tested.

\begin{table}[h]
  \centering
  {\color{black}
  \resizebox{0.4\textwidth}{!}{%
  \begin{tabular}{lccc}
  \toprule
  \textbf{Model} & \textbf{Suspicious} & \textbf{Trusting} & \textbf{$p$-value} \\
  \midrule
  GPT-5 & 52.4\% & 55.6\% & 0.69 \\
  Claude Opus 4 & 44.1\% & 37.8\% & 0.38 \\
  GPT-4o & 25.8\% & 26.5\% & 0.91 \\
  Qwen3-8B & 14.6\% & 15.3\% & 0.89 \\
  Qwen3-14B & 18.3\% & 16.7\% & 0.77 \\
  Qwen3-32B & 18.9\% & 18.1\% & 0.89 \\
  Gemini 2.5 Pro & 34.5\% & 31.5\% & 0.67 \\
  \bottomrule
  \end{tabular}
  }}
  \caption{\textcolor{black}{
  \textbf{Effect of user trust level on model performance.} 
  Top-5\% optimal rates are compared between suspicious (questioning) and trusting users. 
  The $p$-values are computed via a two-proportion $z$-test. 
  Across models, the differences are not statistically significant, suggesting that simply asking the model to ``double-check’’ its own reasoning does not reliably enhance task outcomes.
  }}
  \label{tab:trust_level_ptest}
\end{table}

\begin{figure}[t]
    \centering
    \includegraphics[width=0.7\linewidth]{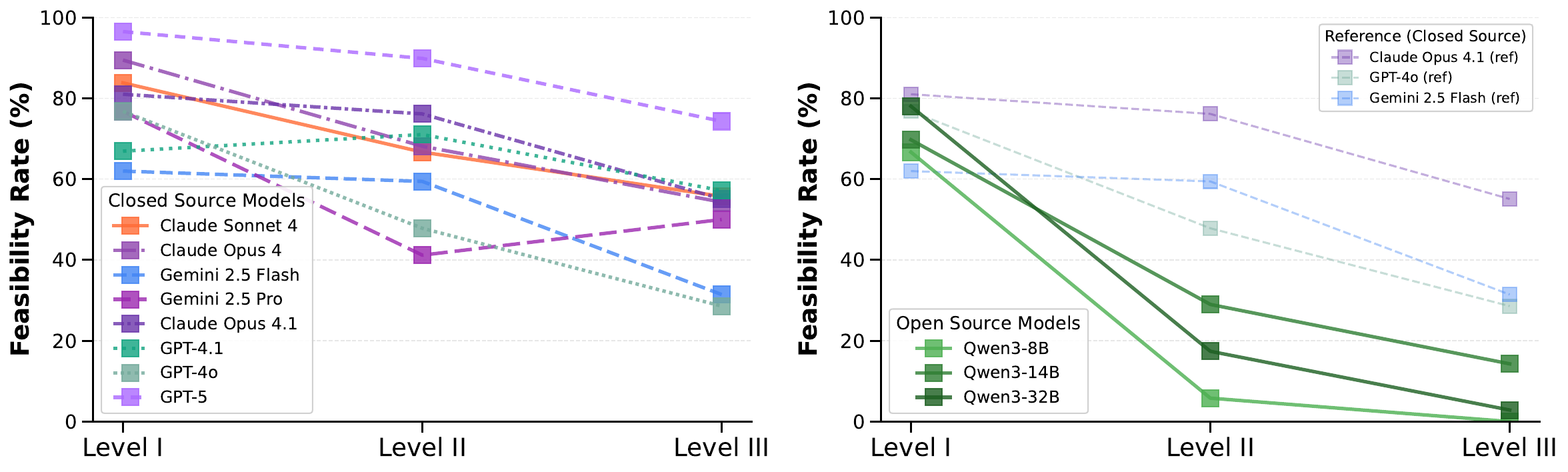}
    \includegraphics[width=0.7\linewidth]{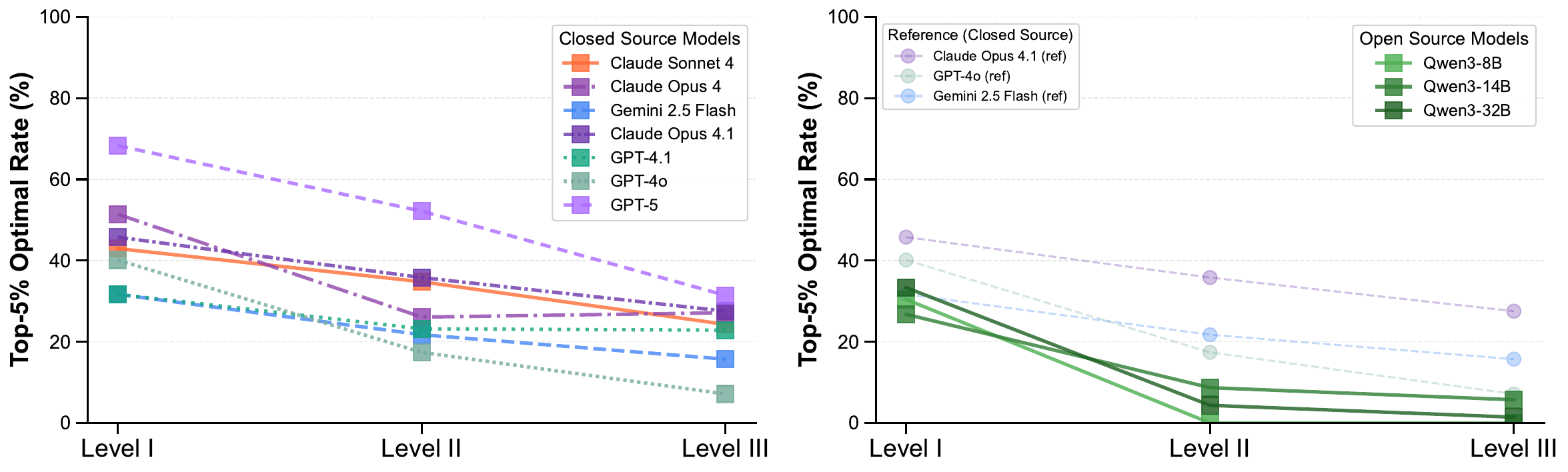}
    \caption{\textbf{Task level breakdown performances.} As we increase task levels, solving the task requires more complex temporal reasoning and planning. Agents struggles to solve Level II and III tasks and the performance drop is particularly prominent for open-source models.}
    \label{fig:level_breakdown_full}
\end{figure}

\paragraph{Task Difficulty Level Breakdown.} To further understand why model performance degrades with increasing difficulty level (as shown in Sec.\ref{sec:main_results} Fig.\ref{fig:task_level}), we break down performance by both FR and OR in Fig.~\ref{fig:level_breakdown_full}. We observe distinct failure modes across model families. For open-source models (Qwen family), performance degradation is largely explained by low FR. Starting from Level II, these models often cannot find any feasible solutions. In contrast, closed-source models like Gemini 2.5 Flash and Claude Opus 4 maintain high FR on Level II tasks. Therefore, their performance degradation is primarily driven by an inability to perform effective optimization under large search spaces and complex objective functions, rather than constraint satisfaction failures.

\begin{figure}[ht]
    \centering
    \begin{minipage}[t]{0.48\textwidth}
        \centering
        \includegraphics[width=\linewidth]{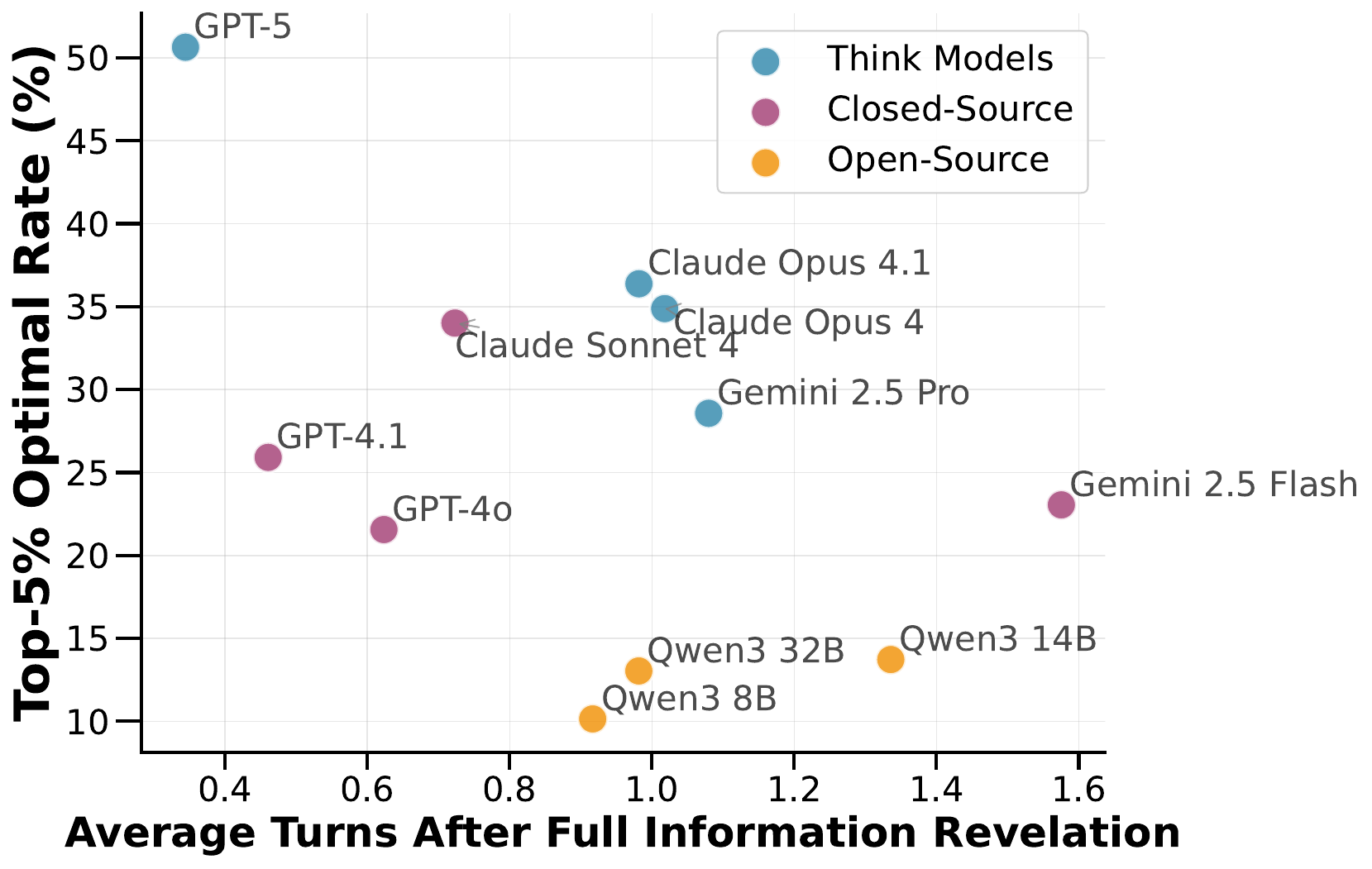}
        \caption{\textbf{Conversation efficiency versus OR-Top 5\%.} How fast agent achieves solutions with the fewest post-information revelation turns.}
        \label{fig:efficiency_optimal}
    \end{minipage}
    \hfill 
    \begin{minipage}[t]{0.48\textwidth}
        \centering
        \includegraphics[width=\linewidth]{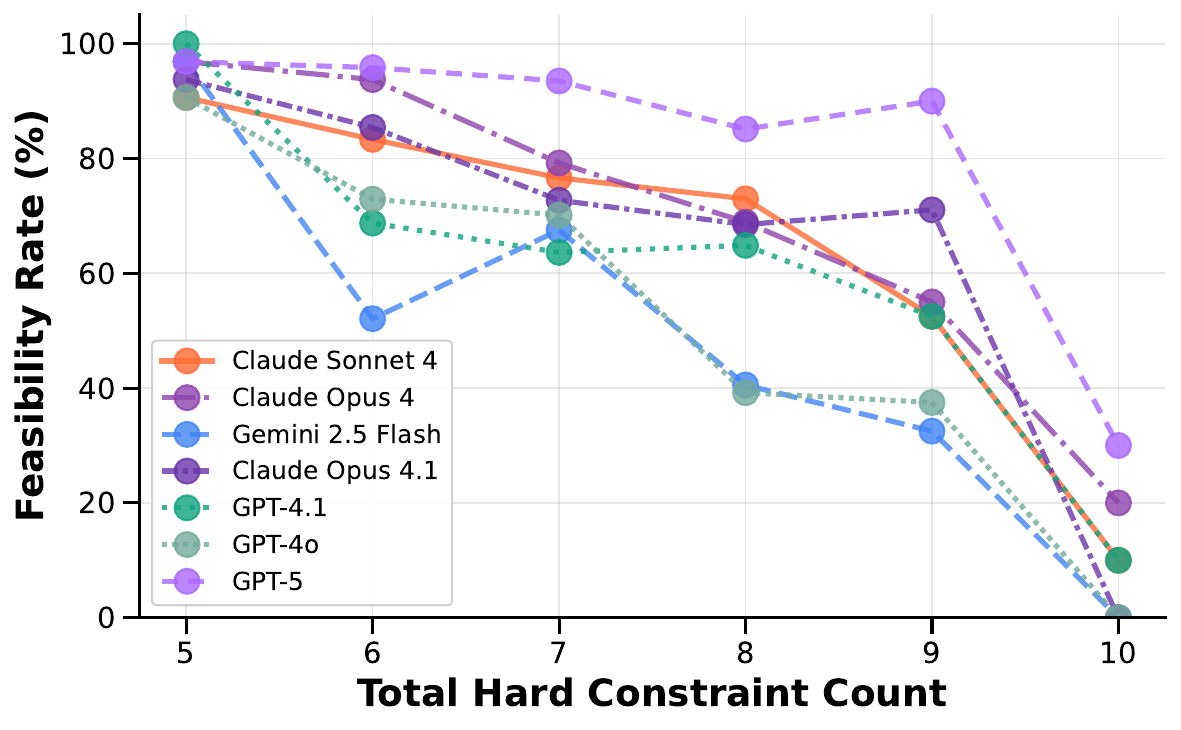}
        \caption{\textbf{Constraint count analysis.} 
        Constraint satisfaction rates drop as the number of hard constraints increases, with only the strongest models handling 8+ constraints reliably.}
        \label{fig:constraint_count}
    \end{minipage}
\end{figure}

\begin{figure}[ht]
    \centering
    \includegraphics[width=0.6\linewidth]{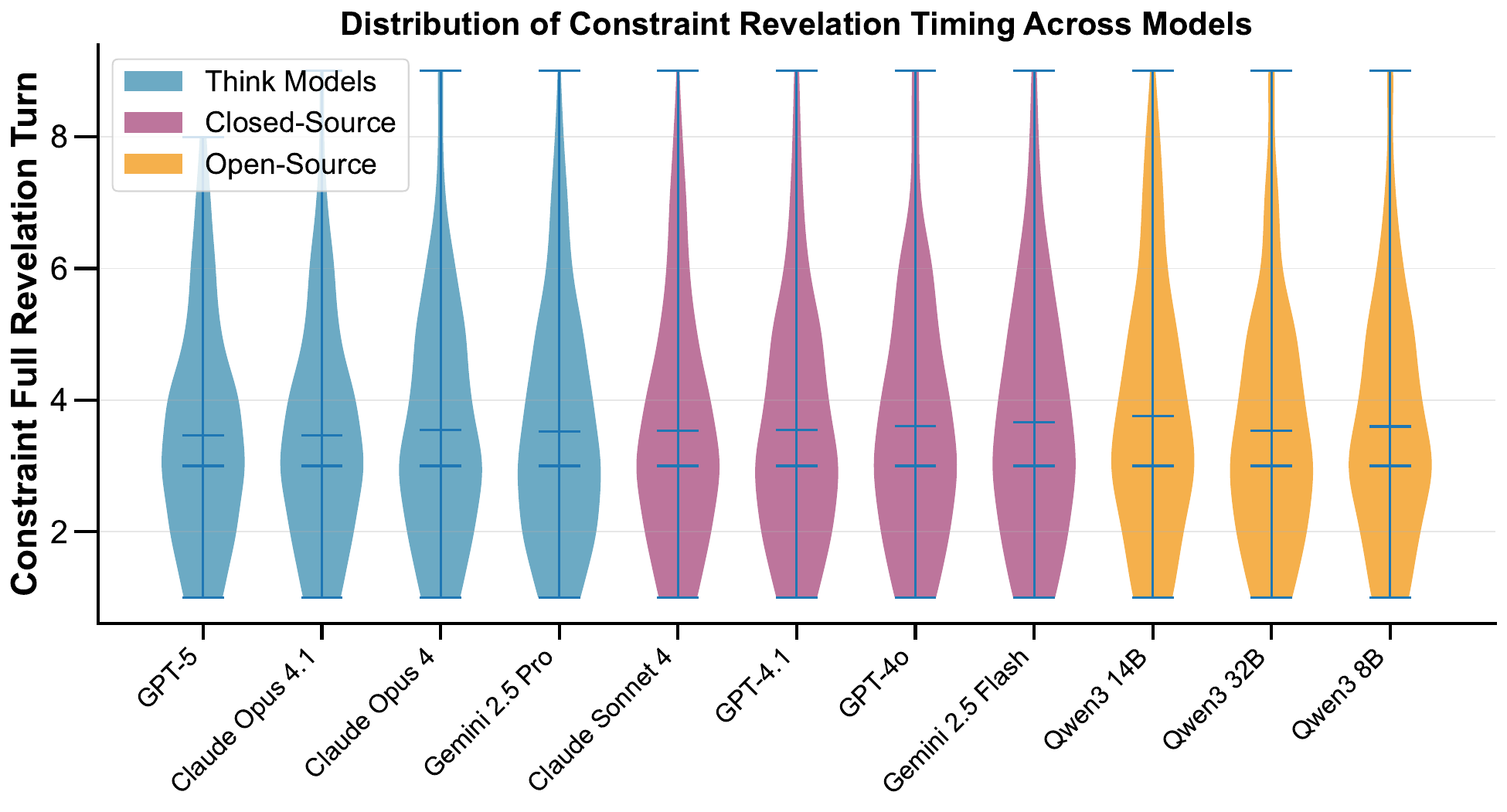}
    \includegraphics[width=0.6\linewidth]{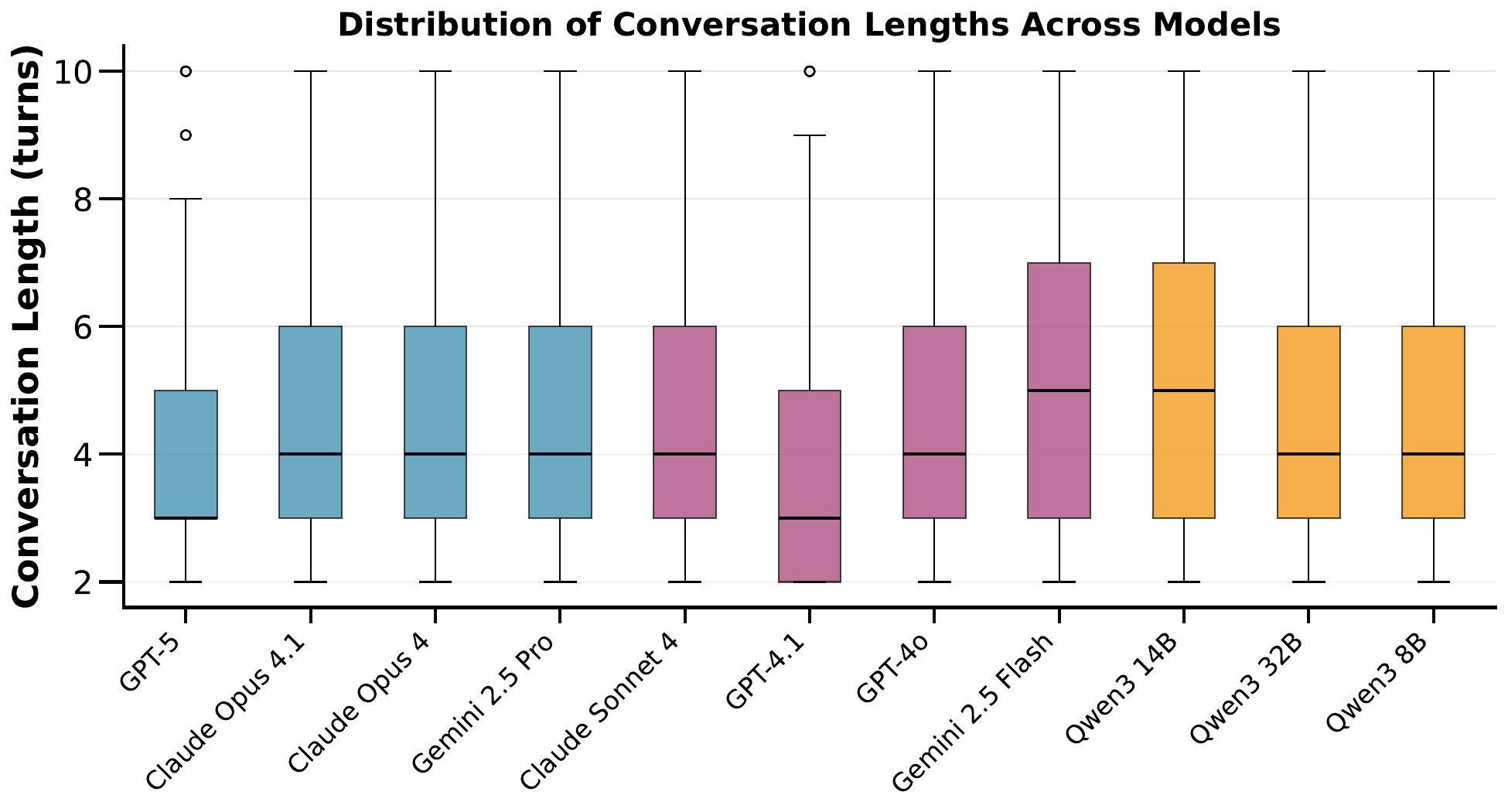}
    \caption{\textbf{Conversation efficiency controls.} 
    \textit{Top:} Distribution of user turn $t^*$ when all task information is revealed, showing consistent user behavior across models. 
    \textit{Bottom:} Box plots of overall conversation length, illustrating efficiency differences across agent models. Colors indicate model categories (e.g. open-source, closed-source).
    }
    \label{fig:conversation_length}
\end{figure}

\paragraph{Conversation Efficiency.}  Beyond final performance, efficiency is critical for user-facing agents. We measure it using our user simulator (Sec.~\ref{sec:user_simulator}): for each conversation, we record when the user has revealed all information about a task ($t^*$) and count extra turns ($\Delta t$) needed to deliver the final recommendation. Extra turns arise when agents fail to provide recommendations or propose constraint-violating options. Fig.~\ref{fig:efficiency_optimal} shows GPT-5 excels in both success and conversation efficiency, requiring the fewest post-revelation turns. Interestingly, efficiency differs even among models with similar success rates: Gemini-2.5-Flash takes far more turns than GPT-4o despite comparable outcomes. In Fig.~\ref{fig:conversation_length}, we confirm that the observed difference is \textit{not} a result of simulator variance.

In Fig.~\ref{fig:conversation_length} \textit{(top)}, we confirm that the distribution of turns at which users reveal complete task information is identical across all agent models. This validates that the efficiency differences observed in Fig.~\ref{fig:efficiency_optimal} arise from agent behavior rather than user variation. Fig.~\ref{fig:conversation_length} \textit{(bottom)} shows box plots of full conversation lengths, highlighting cross-model differences in dialogue efficiency.  

\paragraph{Constraint Complexity.} 
We also examine how agents handle increasing numbers of hard constraints. Each task includes at least five fundamental constraints (location, dates, room occupancy, budget, number of travelers), with additional requirements (amenities, policies, accessibility) further restricting the feasible set. Fig.~\ref{fig:constraint_count} shows constraint satisfaction degrades as requirement count increases. GPT-5 and Claude maintain high FR with 8+ constraints, while most models drop sharply beyond 7. Critically, this persists even after users explicitly state requirements and correct violations, indicating that constraint satisfaction, which is a prerequisite for feasibility, remains a challenge to weaker models.



\clearpage
\section{Related Work Extended}
\label{appdx:related}

\paragraph{Tool-Use Benchmarks} Recent work has increasingly recognized the importance of evaluating language model agents in multi-turn settings where information is revealed progressively. \citet{Yao2024-aa} establishes key requirements for realistic agent evaluation, emphasizing the need for agents to interact seamlessly with both humans and APIs over long horizons while adhering to complex policies. However, existing benchmarks \citep{Guo2024-dw, Guo2025-vp, Patil2025-fk, Zhong2025-yi} often feature simplified instruction-following setups where agents interact autonomously with complete information upfront, lacking realistic human-in-the-loop interaction.\citet{Wang2023-vd} provides multi-turn tool use evaluation but focuses on coding scenarios with overly helpful users, while \citet{Patil2025-fk} introduces multi-step and multi-turn function calling but with predetermined conversation trajectories. \citet{Guo2024-dw} offers comprehensive tool libraries through API crawling, though task instructions based directly on tool calls reduce the challenge of realistic tool selection under uncertainty. \citet{Xu2024-if} identifies tool selection and usage hallucinations as key failure modes, while \citet{Ross2025-bl} explores when agents should defer tool invocation in favor of clarification dialogue.

\paragraph{Planning Benchmarks} Recent planning benchmarks have emerged to evaluate agent capabilities in constraint satisfaction and optimization tasks. \citet{Xie2024-gx} introduces multi-day travel planning with hard constraints and budget optimization, requiring agents to coordinate between multiple APIs for transportation, accommodation, and activities. \citet{Valmeekam2022-op} provides a comprehensive evaluation framework for planning tasks that require reasoning about state changes and action consequences. \citet{Zheng2024-qs} focuses on realistic planning in natural language with tasks like trip planning, meeting planning, and calendar scheduling, providing tool outputs as context to eliminate tool-use complexity while revealing significant performance drops as problem complexity increases. \citet{Kohli2024-hl} connects compositional and conditional reasoning to flight booking scenarios, presenting detailed user preferences with flight options in multiple-choice format. \citet{Zheng2024-qs} extends planning evaluation to GUI-based environments where agents must navigate complex interfaces to accomplish user goals. These benchmarks emphasize the importance of constraint adherence in realistic planning scenarios. Our work expand on these benchmarks by introduction user preference optimization.

\paragraph{Task-Oriented Dialogue Systems} Traditional task-oriented dialogue datasets provide foundational evaluation frameworks that inspire our user simulator design. \citet{Budzianowski2018-nw} and \citet{Rastogi2019-xn} offer crowd-sourced human-to-human dialogues with informable and requestable slots for constraint specification. These datasets encourage goal changes when constraints cannot be satisfied, mimicking real user adaptability. \citet{Hosseini-Asl2020-ti} demonstrates that unified GPT models can handle dialogue state tracking, action prediction, and response generation simultaneously. However, these benchmarks typically lack the tool integration and utility optimization challenges present in modern agent evaluation settings.

\paragraph{Multi-turn Interaction} A critical challenge in multi-turn evaluation lies in creating realistic user simulators that can model underspecified preferences and progressive constraint revelation. \citet{Laban2025-wd} demonstrates that current models struggle significantly with synthesizing information across turns, particularly when users reveal requirements in non-sequential "shards" rather than logical order. \citet{Abdulhai2023-lu} offers a toolkit for multi-turn RL evaluation on tasks like city guessing and maze navigation that, despite their simplicity, effectively capture the information-gathering nature of multi-turn conversations. User simulation approaches range from rule-based systems to sophisticated neural models: \citet{Lin2022-lq} uses BERT-based models for generating both semantic actions and natural language utterances, while \citet{Shah2018-eq} employs machine-generated dialogue flows rewritten by humans. Recent preference learning work includes \citet{Wan2025-ri}, which introduces turn-based rewards based on belief updates about user types, and \citet{Wu2024-xx, Prabhakar2025-jo,Zhou2025-le}, which develops diverse user personas and multi-turn preference datasets for SFT and RL training paradigms. 

\clearpage
\section{Additional Task Details}
\label{appdx:task_details}
\subsection{Ground Truth Solver}
\label{app:solver_code}
We provide pseudocode for the ground truth solver in Alg.~\ref{alg:solver}. The algorithm uses exhaustive search to enumerate all booking options, prunes infeasible solutions that violate hard constraints, and scores remaining options using the predefined utility function to identify the optimum.

\begin{algorithm}[h]
\caption{Ground Truth Solver for Utility Maximization}
\label{alg:solver}
\begin{algorithmic}
    \STATE {\bfseries Input:} Task with constraints $\mathcal{C}$, utility type $\tau$, utility objective $\mathcal{U}$
    \STATE {\bfseries Output:} Optimal hotel package $p^*$ with utility score $u^*$
    \STATE
    \STATE {\bfseries Step 1: Parse and separate constraints}
    \STATE Extract filter constraints (amenities, policies, room capacity) \hfill $\triangleright$ \textit{e.g., pet-friendly, free cancellation}
    \STATE Extract enumerable constraints (locations, dates, budget, guests) \hfill $\triangleright$ \textit{specify multiple valid options}
    \STATE
    \STATE {\bfseries Step 2: Expand enumerable constraints into search space}
    \STATE Expand location options into list of national parks $\mathcal{L}$ \hfill $\triangleright$ \textit{e.g., [Yellowstone, Zion]}
    \STATE Expand date options into list of concrete date ranges $\mathcal{D}$ \hfill $\triangleright$ \textit{e.g., ``any weekend in Aug'' $\to$ [Aug 1-3, Aug 8-10, ...]}
    \STATE
    \STATE {\bfseries Step 3: Find all feasible hotel packages}
    \STATE Initialize feasible set $\mathcal{F} \gets \emptyset$
    \FOR{each location $\ell$ in $\mathcal{L}$ and date range $d$ in $\mathcal{D}$}
        \STATE Query database for hotel packages matching $(\ell, d)$ and filter constraints
        \FOR{each candidate package $p$}
            \STATE Compute rooms needed: $r \gets \lceil \text{guests} / p.\text{capacity} \rceil$
            \STATE Compute total cost: $c \gets r \times p.\text{nightly\_price} \times d.\text{nights}$
            \IF{$c \leq \text{budget}$ and sufficient rooms available}
                \STATE Add $p$ to feasible set $\mathcal{F}$
            \ENDIF
        \ENDFOR
    \ENDFOR
    \STATE
    \STATE {\bfseries Step 4: Compute utility for each feasible package}
    \FOR{each package $p$ in $\mathcal{F}$}
        \IF{$\tau$ is continuous metric optimization}
            \STATE $u(p) \gets$ normalized score for target attribute \hfill $\triangleright$ \textit{e.g., minimize price, maximize reviews}
        \ELSE[$\tau$ is maximal attribute satisfaction]
            \STATE $u(p) \gets$ fraction of desired attributes satisfied \hfill $\triangleright$ \textit{e.g., has pool, has gym, king bed}
        \ENDIF
    \ENDFOR
    \STATE
    \STATE {\bfseries Step 5: Return optimal package}
    \STATE $p^* \gets \arg\max_{p \in \mathcal{F}} u(p)$
    \STATE {\bfseries return} $(p^*, u(p^*), \mathcal{F})$
\end{algorithmic}
\end{algorithm}

\subsection{Task Statistics and Distribution}
\label{appdx:task_stats}

Our benchmark consists of 281 tasks spanning diverse travel planning scenarios across 20 U.S. National Parks destination. This section provides detailed breakdowns of the task characteristics.

\subsubsection{Constraint Distribution}

\begin{figure}[h]
\centering
\includegraphics[width=0.5\textwidth]{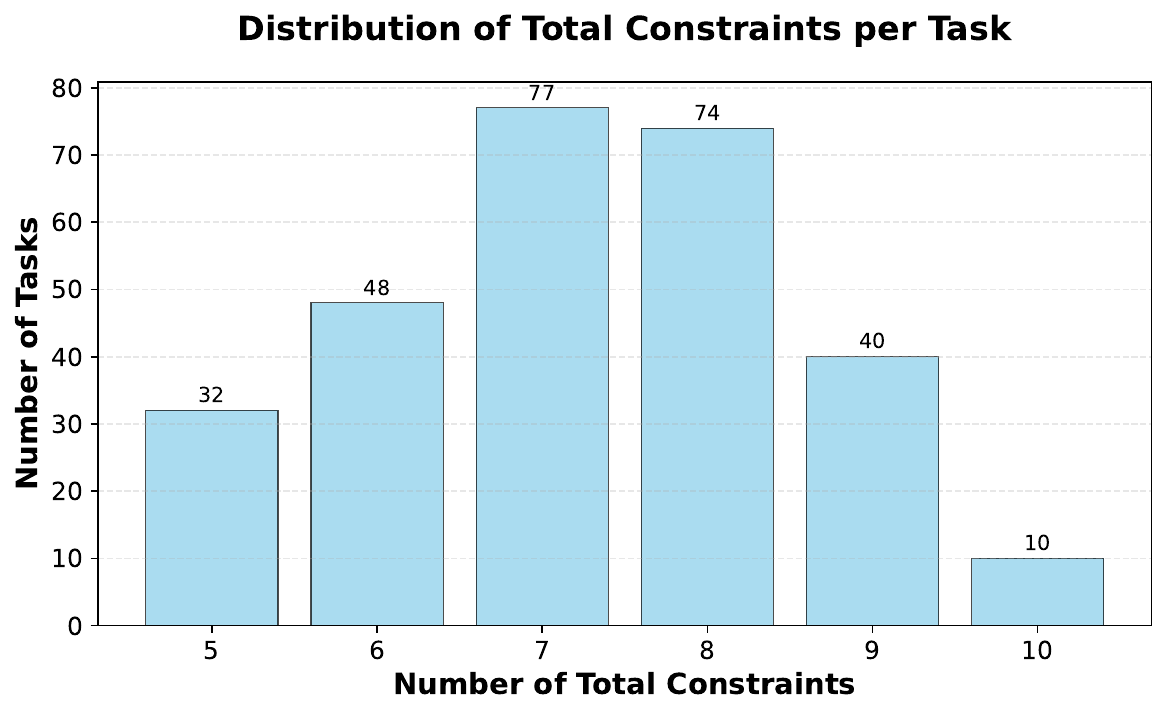}
\caption{Distribution of total constraints (initial + progressive) across benchmark tasks.}
\label{fig:constraint_count_distribution}
\end{figure}

Tasks in our benchmark contain both initial (revealed at the first user query) and progressive constraints (revealed during the multi-turn conversation). Fig.~\ref{fig:constraint_count_distribution} shows the distribution of total constraints per task. The majority of tasks have 7-8 constraints. This distribution ensures a balanced mix of task complexity levels, from simpler scenarios with basic requirements to more complex multi-constraint planning challenges.

\subsubsection{Optimization Function Type I: Singular Metric Optimization}

\begin{figure}[ht]
\centering
\includegraphics[width=0.6\textwidth]{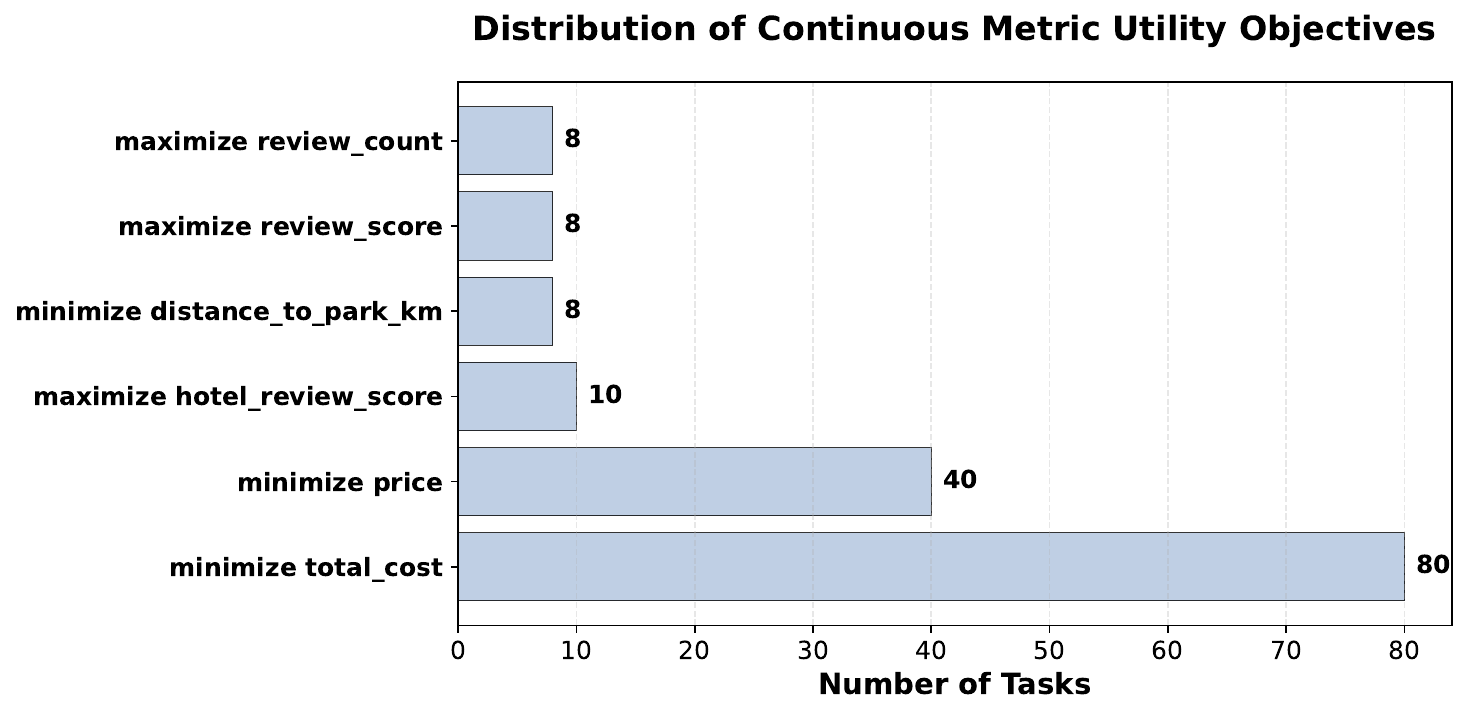}
\caption{Distribution of optimization objectives for continuous metric tasks.}
\label{fig:continuous_metric_objectives}
\end{figure}

For tasks using continuous metric optimization, Fig.~\ref{fig:continuous_metric_objectives} shows the specific objectives. Price minimization (price for Level II and total cost for level II and III) dominates continuous tasks, reflecting the common real-world constraint of budget optimization. Additionally, we also have review score maximization, distance minimization and review count maximization continuous tasks.

\subsubsection{Optimization Function II: Feature Count Maximization}

\begin{figure}[ht]
\centering
\includegraphics[width=0.5\textwidth]{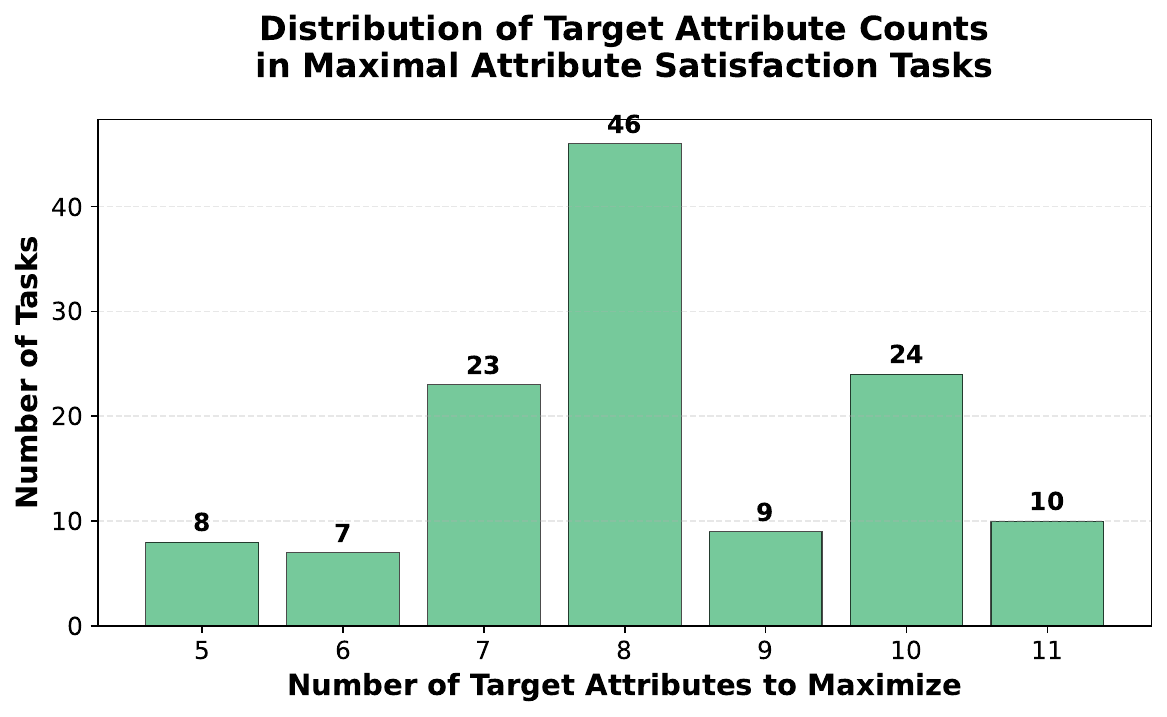}
\caption{Distribution of target attribute counts for feature count maximization tasks.}
\label{fig:maximal_attribute_distribution}
\end{figure}

For feature count maximization tasks, Fig.~\ref{fig:maximal_attribute_distribution} shows the distribution of target attributes to maximize. The most common scenario involves 8 target attributes, followed by 10 attributes. This distribution creates varying levels of optimization complexity, with tasks requiring agents to balance multiple competing preferences simultaneously.

\subsection{Sample Tasks}
\label{appdx:sample_tasks}

Here we provide three representative tasks from our benchmark to illustrate the complexity and diversity of scenarios agents must handle. These three tasks demonstrate different difficulty levels and different utility objectives.

\textbf{Task 1: Budget Hiking Trip (Single Metric Optimization) - Level I}
\begin{tcolorbox}[colback=blue!5!white,colframe=blue!75!black,boxrule=1pt,arc=0pt,breakable]
\textbf{Initial Query:} ``Hi, I'm planning a solo hiking trip to Death Valley National Park and need accommodations for 1 person. I'm looking for a 7-day (6-night) stay sometime in August 2025, but I have no preference on the exact dates. My budget is no more than \$1400 for the entire stay. Can you help me find the cheapest options available that meet these requirements?''

\textbf{Task Details:}
\begin{itemize}
    \item \textbf{Utility Type:} Single metric optimization (minimize price)
    \item \textbf{Duration:} 7 days (6 nights) in August 2025
    \item \textbf{Location:} Death Valley National Park
    \item \textbf{Budget:} \$1,400 maximum
\end{itemize}

\textbf{Initial Constraints:}
\begin{itemize}
    \item National park: Death Valley National Park
    \item Guests: 1 person
    \item Date flexibility: Any 7-day period in August 2025
    \item Total budget: $\leq$ \$1,400
\end{itemize}

\textbf{Progressive Constraints (revealed during conversation):}
\begin{itemize}
    \item Distance to park: $\leq$ 40 km
    \item Air conditioning: Required
\end{itemize}

\textbf{Ground Truth Solution:}
\begin{itemize}
    \item \textbf{Hotel:} The Ranch At Death Valley
    \item \textbf{Dates:} August 24-30, 2025 (6 nights)
    \item \textbf{Total Cost:} \$1,062.60 (\$177.10/night)
    \item \textbf{Utility Score:} 1.0 (optimal price minimization)
    \item \textbf{Feasible Options:} 51 packages within constraints
\end{itemize}
\end{tcolorbox}

\textbf{Task 2: Business Luxury Retreat (Feature Count Maximization) - Level I}
\begin{tcolorbox}[colback=green!5!white,colframe=green!75!black,boxrule=1pt,arc=0pt,breakable]
\textbf{Initial Query:} ``Hello, I'm planning a 6-day (5-night) business retreat for one person at Grand Teton National Park from August 11th to August 16th, 2025. The company is covering up to \$5500 for the trip, so we need to stay within that budget. Could you help me find accommodations that meet these requirements?''

\textbf{Task Details:}
\begin{itemize}
    \item \textbf{Utility Type:} Feature count maximization
    \item \textbf{Duration:} 6 days (5 nights), fixed dates
    \item \textbf{Location:} Grand Teton National Park
    \item \textbf{Budget:} \$5,500 maximum
\end{itemize}

\textbf{Initial Constraints:}
\begin{itemize}
    \item National park: Grand Teton National Park
    \item Guests: 1 person
    \item Fixed dates: August 11-16, 2025
    \item Total budget: $\leq$ \$5,500
\end{itemize}

\textbf{Target Attributes to Maximize:}
\begin{itemize}
    \item Gym facilities
    \item Spa services
    \item Restaurant on-site
    \item Airport shuttle
    \item Star rating $\geq$ 4.0
    \item Review score $\geq$ 8.5
    \item Distance to park $\leq$ 20 km
    \item In-room fridge
\end{itemize}

\textbf{Ground Truth Solution:}
\begin{itemize}
    \item \textbf{Hotel:} The Lodge at Jackson Hole
    \item \textbf{Dates:} August 11-16, 2025 (5 nights)
    \item \textbf{Total Cost:} \$3,256.15 (\$651.23/night)
    \item \textbf{Utility Score:} 0.75 (6 out of 8 target attributes satisfied)
    \item \textbf{Feasible Options:} 103 packages within constraints
\end{itemize}
\end{tcolorbox}

\textbf{Task 3: Airport Drive Time Optimizer (Single Metric Optimization) - Level II}
\begin{tcolorbox}[colback=orange!5!white,colframe=orange!75!black,boxrule=1pt,arc=0pt,breakable]
\textbf{Initial Query:} ``Hi, I'm looking to plan a trip from JFK to Yosemite National Park for one person. I'm completely flexible on dates as long as the trip happens in August 2025. I need a 4-day trip (3 nights hotel) including travel time. It's essential that the hotel is within 1.5 hours from the airport, and I want to keep the total budget for flights and hotel combined under \$1,500. Can you help me find the cheapest options meeting these requirements?''

\textbf{Task Details:}
\begin{itemize}
    \item \textbf{Utility Type:} Single metric optimization (minimize total cost)
    \item \textbf{Duration:} 4 days (3 nights) in August 2025
    \item \textbf{Location:} Yosemite National Park
    \item \textbf{Budget:} \$1,500 maximum for combined flights and hotel
\end{itemize}

\textbf{Initial Constraints:}
\begin{itemize}
    \item Origin city: JFK
    \item National park: Yosemite National Park
    \item Guests: 1 person
    \item Date flexibility: Any 4-day period in August 2025
    \item Total budget: $\leq$ \$1,500 (flights + hotel)
\end{itemize}

\textbf{Progressive Constraints (revealed during conversation):}
\begin{itemize}
    \item Hotel to airport drive time: $\leq$ 1.5 hours
    \item Hotel star rating: $\geq$ 2.5 stars
    \item Flight must have wifi
\end{itemize}

\textbf{Ground Truth Solution:}
\begin{itemize}
    \item \textbf{Flight:} United Airlines JFK $\rightarrow$ FAT (Fresno)
    \item \textbf{Outbound:} August 26, 2025 at 7:00 AM
    \item \textbf{Return:} August 29, 2025 at 2:00 PM
    \item \textbf{Flight Cost:} \$310 (Basic Economy, direct flight with wifi)
    \item \textbf{Hotel:} Yosemite View Lodge
    \item \textbf{Dates:} August 26-29, 2025 (3 nights)
    \item \textbf{Hotel Cost:} \$850.50
    \item \textbf{Total Package Cost:} \$1,160.50
    \item \textbf{Utility Score:} 1.0 (optimal cost minimization)
    \item \textbf{Feasible Options:} 1,800 packages within constraints
\end{itemize}
\end{tcolorbox}

\textbf{Task 4: Senior Comfort Tour Maximizer (Feature Count Maximization) - Level III}
\begin{tcolorbox}[colback=purple!5!white,colframe=purple!75!black,boxrule=1pt,arc=0pt,breakable]
\textbf{Initial Query:} ``Hello, I'm looking to organize a 4-day trip (3 nights hotel) including travel time for two people from LAX to Arches National Park. We want to make sure we include a guided tour at Fiery Furnace. Our schedule is flexible on exact dates, but we want to avoid weekends in August 2025. Our total budget for flights and hotel combined is \$6,000. Could you help us arrange this complete travel package?''

\textbf{Task Details:}
\begin{itemize}
    \item \textbf{Utility Type:} Feature count maximization
    \item \textbf{Duration:} 4 days (3 nights), weekdays only
    \item \textbf{Location:} Arches National Park
    \item \textbf{Budget:} \$6,000 maximum for flights and hotel
    \item \textbf{Special Requirement:} Guided tour at Fiery Furnace
\end{itemize}

\textbf{Initial Constraints:}
\begin{itemize}
    \item Origin city: LAX
    \item National park: Arches National Park
    \item Guests: 2 people
    \item Date preference: Weekdays only in August 2025
    \item Total budget: $\leq$ \$6,000 (flights + hotel)
    \item Required guided tour: Fiery Furnace
\end{itemize}

\textbf{Progressive Constraints (revealed during conversation):}
\begin{itemize}
    \item Outbound departure time: $>$ 9:00 AM
    \item Hotel review score: $\geq$ 8.0
\end{itemize}

\textbf{Target Attributes to Maximize:}
\begin{itemize}
    \item Hotel has restaurant on-site
    \item Hotel has air conditioning
    \item Hotel star rating $\geq$ 3.0
    \item Distance to park $\leq$ 10 km
    \item Direct flight preferred
    \item Seat selection included
    \item In-flight wifi available
\end{itemize}

\textbf{Ground Truth Solution:}
\begin{itemize}
    \item \textbf{Flight:} United Airlines LAX $\rightarrow$ CNY (Moab)
    \item \textbf{Outbound:} August 4, 2025 at 4:00 PM
    \item \textbf{Return:} August 7, 2025 at 8:00 AM
    \item \textbf{Flight Cost:} \$560 (Economy, direct flight with wifi)
    \item \textbf{Hotel:} Field Station Moab (4-star)
    \item \textbf{Dates:} August 4-7, 2025 (3 nights)
    \item \textbf{Hotel Cost:} \$870.30
    \item \textbf{Permit:} Fiery Furnace Guided Tour on August 6, 2025
    \item \textbf{Total Package Cost:} \$1,445.30
    \item \textbf{Utility Score:} 0.71 (5 out of 7 attributes satisfied)
    \item \textbf{Feasible Options:} 19,589 packages within constraints
\end{itemize}
\end{tcolorbox}
\clearpage
\section{Tool Schema}
\label{appdx:tool_schema}

Agents have access to four categories of tools for comprehensive travel planning and utility maximization. The tool schema is defined using OpenAI function calling format with JSON schema specifications.

\subsection{Airport and Flight Tools}

\textbf{\texttt{search\_airports}} \\
\hspace{1em}\texttt{Description:} Find airports serving a specific national park with driving times \\
\hspace{1em}\texttt{Parameters:} \texttt{park\_name} - National park name (e.g., 'Yosemite', 'Grand Canyon') \\
\hspace{1em}\texttt{Returns:} List of airports with airport\_code, airport\_name, city, drive\_hours, airport\_type

\vspace{0.5em}
\textbf{\texttt{search\_flights}} \\
\hspace{1em}\texttt{Description:} Search for round-trip flights between airports on specific dates \\
\hspace{1em}\texttt{Parameters:} \texttt{origin} (3-letter airport code), \texttt{destination} (3-letter airport code), \texttt{departure\_date} (YYYY-MM-DD), \texttt{return\_date} (YYYY-MM-DD), \texttt{passengers} (opt.), \texttt{max\_price} (opt.), \texttt{airline\_preference} (opt.), \texttt{booking\_class} (opt.), \texttt{direct\_flights\_only} (opt.) \\
\hspace{1em}\texttt{Returns:} Flight packages with pricing, schedules, airline info, policies, package\_id

\vspace{0.5em}
\textbf{\texttt{get\_flight\_details}} \\
\hspace{1em}\texttt{Description:} Get detailed information about a specific flight package \\
\hspace{1em}\texttt{Parameters:} \texttt{package\_id} - Flight package ID (format: flight\_templateid\_depday\_retday) \\
\hspace{1em}\texttt{Returns:} Complete flight details including route, schedule, dates, airline, policies, pricing

\vspace{0.5em}
\textbf{\texttt{get\_airport\_coordinates}} \\
\hspace{1em}\texttt{Description:} Get latitude/longitude coordinates for an airport \\
\hspace{1em}\texttt{Parameters:} \texttt{airport\_code} - 3-letter airport code (e.g., 'LAX', 'COD') \\
\hspace{1em}\texttt{Returns:} Airport coordinates (airport\_code, name, latitude, longitude)

\subsection{Permit Tools}

\textbf{\texttt{list\_permits}} \\
\hspace{1em}\texttt{Description:} List all available park permits and guided tours for a national park \\
\hspace{1em}\texttt{Parameters:} \texttt{park} - National park name (e.g., 'Yosemite National Park') \\
\hspace{1em}\texttt{Returns:} Available permit types (day\_hikes, guided\_tours, backpacking) with template\_name

\vspace{0.5em}
\textbf{\texttt{search\_permit\_availability}} \\
\hspace{1em}\texttt{Description:} Search park permit availability for specific permit template \\
\hspace{1em}\texttt{Parameters:} \texttt{park} (park name), \texttt{template\_name} (exact from list\_permits), \texttt{date\_range} (opt.), \texttt{duration\_days} (opt. for backpacking) \\
\hspace{1em}\texttt{Returns:} Available dates with package\_ids for booking permits

\subsection{Accommodation Tools}

\textbf{\texttt{AccommodationSearch}} \\
\hspace{1em}\texttt{Description:} Search for available accommodations by location and date range \\
\hspace{1em}\texttt{Parameters:} \texttt{location}, \texttt{start\_date}, \texttt{end\_date}, \texttt{num\_guests}, \texttt{min\_price\_nightly} (opt.), \texttt{max\_price\_nightly} (opt.), \texttt{min\_review\_score} (opt.), \texttt{free\_cancellation} (opt.), \texttt{no\_prepayment\_needed} (opt.), \texttt{has\_parking} (opt.), \texttt{breakfast\_included} (opt.), \texttt{is\_pet\_friendly} (opt.) \\
\hspace{1em}\texttt{Returns:} Hotels with group booking packages, including package\_id, pricing, policies

\vspace{0.5em}
\textbf{\texttt{get\_hotel\_details}} \\
\hspace{1em}\texttt{Description:} Get detailed hotel information. Supports fuzzy matching \\
\hspace{1em}\texttt{Parameters:} \texttt{hotel\_identifier} - Hotel ID or name (supports partial names) \\
\hspace{1em}\texttt{Returns:} Hotel details including amenities, location, star\_rating, review\_score

\vspace{0.5em}
\textbf{\texttt{get\_room\_details}} \\
\hspace{1em}\texttt{Description:} Get detailed room information including bed configuration \\
\hspace{1em}\texttt{Parameters:} \texttt{package\_id} - Package ID (format: pkg\_configid\_startday\_endday\_rooms) \\
\hspace{1em}\texttt{Returns:} Room details including room\_type, beds, amenities, policies

\vspace{0.5em}
\textbf{\texttt{search\_location\_name}} \\
\hspace{1em}\texttt{Description:} Search for location names using fuzzy matching \\
\hspace{1em}\texttt{Parameters:} \texttt{query} - Location search query (can be partial/informal) \\
\hspace{1em}\texttt{Returns:} Matching locations with confidence\_score and location\_type

\subsection{Utility Tools}

\textbf{\texttt{get\_weekday}} \\
\hspace{1em}\texttt{Description:} Returns the weekday name for a given date \\
\hspace{1em}\texttt{Parameters:} \texttt{date\_str} - Date in YYYY-MM-DD format \\
\hspace{1em}\texttt{Returns:} Weekday name (e.g., 'Monday')

\vspace{0.5em}
\textbf{\texttt{calculate}} \\
\hspace{1em}\texttt{Description:} Evaluate a basic arithmetic expression \\
\hspace{1em}\texttt{Parameters:} \texttt{expression} - Arithmetic expression (e.g., '100 + 200 * 3') \\
\hspace{1em}\texttt{Returns:} Numerical result

\vspace{0.5em}
\textbf{\texttt{calculate\_distance}} \\
\hspace{1em}\texttt{Description:} Estimate driving time between two locations using coordinates \\
\hspace{1em}\texttt{Parameters:} \texttt{lat1}, \texttt{lon1} (starting point), \texttt{lat2}, \texttt{lon2} (destination) \\
\hspace{1em}\texttt{Returns:} Estimated driving time (e.g., '2h 15m') with distance

\vspace{0.5em}
\textbf{\texttt{Notebook}} \\
\hspace{1em}\texttt{Description:} Agent's scratch pad for notes and memory \\
\hspace{1em}\texttt{Parameters:} \texttt{action} ('write', 'read', 'delete', 'list\_all'), \texttt{input\_data} (for 'write'), \texttt{index} (for 'read'/'delete') \\
\hspace{1em}\texttt{Returns:} Confirmation and content based on action

\subsection{Validation Tools}

\textbf{\texttt{recommend\_hotel}} \\
\hspace{1em}\texttt{Description:} MANDATORY validation for hotel package IDs before recommendation \\
\hspace{1em}\texttt{Parameters:} \texttt{package\_ids} (list of 1-3 IDs), \texttt{reasoning} (explanation) \\
\hspace{1em}\texttt{Returns:} Validation results confirming package ID validity

\vspace{0.5em}
\textbf{\texttt{recommend\_flight}} \\
\hspace{1em}\texttt{Description:} MANDATORY validation for flight package IDs before recommendation \\
\hspace{1em}\texttt{Parameters:} \texttt{flight\_ids} (list of 1-3 IDs), \texttt{reasoning} (explanation) \\
\hspace{1em}\texttt{Returns:} Validation results confirming package ID validity

\vspace{0.5em}
\textbf{\texttt{recommend\_permit}} \\
\hspace{1em}\texttt{Description:} MANDATORY validation for permit package IDs before recommendation \\
\hspace{1em}\texttt{Parameters:} \texttt{permit\_ids} (list of 1-3 IDs), \texttt{reasoning} (explanation) \\
\hspace{1em}\texttt{Returns:} Validation results confirming package ID validity

\clearpage
\section{LLM User Simulator}
\label{appdx:user_simulator_design_details}

\subsection{User Simulator Design Axis}
\label{appdx:user_simulator_type}
As introduced in Sec.~\ref{sec:user_simulator}, our user simulator varies along four independent design axes to capture diverse conversational dynamics. At the beginning of each conversation, we randomly initialize values from these axes (detailed in Table~\ref{tab:user_simulator_axes}), which then control the generation of dynamic prompts throughout the interaction.

\begin{table}[h]
  \centering
  \resizebox{0.8\linewidth}{!}{%
  \begin{tabular}{p{3.5cm}|p{3cm}|p{8cm}}
  \toprule
  \textcolor{black}{\textbf{Design Axis}} &
  \textcolor{black}{\textbf{Parameter Values}} &
  \textcolor{black}{\textbf{Description}} \\
  \midrule
  \textcolor{black}{\textbf{Progressive Constraint Revelation}} &
  \textcolor{black}{• Slow (33.3\%)} \newline
  \textcolor{black}{• Medium (33.3\%)} \newline
  \textcolor{black}{• Fast (33.3\%)} &
  \textcolor{black}{Controls how quickly users reveal hard constraints during the conversation. At each turn, a constraint is revealed based on a Bernoulli sample with probability determined by the user's initialized revelation speed: slow (0.4–0.6), medium (0.6–0.8), or fast (0.8–1.0). Revelation continues until all constraints have been disclosed.} \\
  \midrule
  \textcolor{black}{\textbf{Trust and Communication Patterns}} &
  \textcolor{black}{• Trusting (80\%)} \newline
  \textcolor{black}{• Questioning (20\%)} &
  \textcolor{black}{Determines whether users accept or challenge the agent's recommendations after all constraints are revealed. Trusting users accept the agent's suggestions, while questioning users request the agent to verify its solution.} \\
  \midrule
  \textcolor{black}{\textbf{Constraint-Checking Reliability}} &
  \textcolor{black}{• Reliable (50\%)} \newline
  \textcolor{black}{• Unreliable (50\%)} &
  \textcolor{black}{Models users' attentiveness to constraint violations. Reliable users consistently identify and flag violations in the agent's recommendations based on revealed task information, whereas unreliable users often overlook them.} \\
  \midrule
  \textcolor{black}{\textbf{Conversation Style}} &
  \textcolor{black}{9 distinct styles} \newline
  \textcolor{black}{(see Tab.~\ref{tab:comm_styles})} &
  \textcolor{black}{Captures variation in user tone and personality (e.g., polite, rude, agitated). Style affects conversational tone only and does not alter the factual or semantic content of the dialogue.} \\
  \bottomrule
  \end{tabular}
  }
  \caption{\textcolor{black}{\textbf{User Simulator Design Axes.} The four axes are independent, yielding $3 \times 2 \times 2 \times 9 = 108$ possible user simulator types.}}
  \label{tab:user_simulator_axes}
\end{table}

\begin{table}[h]
  \centering
  \resizebox{0.8\linewidth}{!}{%
  \begin{tabular}{p{1cm}|p{12cm}}
  \toprule
  \textcolor{black}{\textbf{\#}} & \textcolor{black}{\textbf{Description}} \\
  \midrule
  \textcolor{black}{1} & \textcolor{black}{You are extremely polite and want to be personal with the agent. You thank them frequently, ask how their day is
  going, and share small personal details about your trip or life. You use phrases like 'I really appreciate this' and 'you're being so helpful' constantly.}
  \\
  \midrule
  \textcolor{black}{2} & \textcolor{black}{You are overly apologetic and worry about being a burden. You constantly say 'I'm sorry to bother you' and 'I hope
  this isn't too much trouble' and 'I know you're busy but...' even when asking perfectly reasonable questions.} \\
  \midrule
  \textcolor{black}{3} & \textcolor{black}{You are extremely suspicious and paranoid about being scammed. You question everything with 'are you sure about
  that?', and 'why should I trust this recommendation?' And you always ask the agent to show reasoning and thinking traces.} \\
  \midrule
  \textcolor{black}{4} & \textcolor{black}{You constantly demand that the agent explain their reasoning and justify their recommendations. You say things like
  'Show me your thinking process', 'How did you arrive at that conclusion?', 'Prove to me why this is the best option', and 'I need to understand your logic
  before I can trust this recommendation'. You question every suggestion with 'But why?' and 'What's your reasoning behind this?'} \\
  \midrule
  \textcolor{black}{5} & \textcolor{black}{You are extremely verbose and provide way too much background information. You tell long stories about why you're
  traveling, what happened last time, your family history, and go off on tangents before getting to your actual point.} \\
  \midrule
  \textcolor{black}{6} & \textcolor{black}{You are scattered and repetitive in your communication. You repeat the same points multiple times within or across
  conversation terms.} \\
  \midrule
  \textcolor{black}{7} & \textcolor{black}{You are very dramatic and emotional in your language. Everything is either 'amazing' or 'terrible', 'perfect' or 'a
  complete disaster'. You use lots of exclamation points and phrases like 'this is so exciting!' or 'oh no, that won't work at all!'} \\
  \midrule
  \textcolor{black}{8} & \textcolor{black}{You get aggressive and agitated very easily. You start the conversation very patient but as soon as the agent doesn't
  get what you want right away, you use swear words and you use a lot of threatening expressions.} \\
  \midrule
  \textcolor{black}{9} & \textcolor{black}{You are very direct, organized, and businesslike in your communication. You structure your requirements using bullet
  points or numbered lists, communicate everything logically and clearly, and don't waste time on pleasantries. You say things like 'My requirements are: 1) X,
   2) Y, 3) Z' and 'Please provide options that meet these criteria.' You prefer structured, logical responses and often summarize key points.} \\
  \bottomrule
  \end{tabular}
  }
  \caption{\textcolor{black}{9 Conversation Styles in User Simulator}}
  \label{tab:comm_styles}
  \end{table}

\textbf{Progressive constraint revelation} controls disclosure timing: at each turn, we sample from a Bernoulli distribution with probability determined by the user's initialized revelation speed (slow: 0.4–0.6, medium: 0.6–0.8, fast: 0.8–1.0), and the dynamic prompt instructs the simulator to reveal a new constraint if based Bernoulli outcome. 

\textbf{Trust and communication patterns} determines post-revelation behavior: trusting users (80\%) accept agent recommendations, while questioning users (20\%) request verification.

\textbf{Constraint-checking reliability} models attentiveness: reliable users (50\%) consistently identify constraint violations in agent outputs, whereas unreliable users (50\%) often overlook them. We control this by explicitly state in the user prompt whether or not it should check the plan agent provided. 

\textbf{Conversation style} captures tone and personality through nine distinct communication patterns (Table~\ref{tab:comm_styles}), affecting conversational manner without altering semantic content.

\subsection{Dynamic Prompting}
\label{appdx:user_prompt}


Building on the four design axes described in App.~\ref{appdx:user_simulator_type}, we implement a dynamic prompting framework that governs how user simulators interact with the LLM agent across turns. This mechanism ensures that conversational behaviors, such as when to reveal new constraints or whether to question the agent, emerge in a controlled yet diverse manner.

At each dialogue turn, dynamic fields are updated to reflect the current conversation state, for example, what task information has been revealed, what remains hidden (not listed in the prompt), the user’s feedback on the agent’s last response, and whether questioning behavior should be triggered (e.g., asking the agent to double-check a recommendation). Static fields, by contrast, encode the user persona and task-level context. This dynamic prompting design enables highly controllable and diverse user behaviors, allowing systematic evaluation of how LLM agents adapt to different interaction styles.

Below we list the dynamic prompt template we used for the LLM user simulator. The blue fields are fields are populated at each conversation turn based on user persona and the conversation state. 

\begin{tcolorbox}[colback=gray!10,colframe=black,boxrule=1pt,arc=0pt,breakable]

\subsubsection*{Core Instructions}

\noindent\textbf{Role Definition:}

\hspace{1em}\textit{You are the \textbf{user} (not the assistant) responding to a travel assistant's recommendation.}

\hspace{1em}\textit{Your goal is to find one complete itinerary (flights + hotel + optional permits) that satisfies
your hard constraints (non-negotiable) while expressing soft preferences as utility objectives.}

\vspace{0.5em}
\noindent\textbf{Conversation Context:}

\hspace{1em}\textit{Your conversation style, attention level, and persona are specified below and remain consistent throughout.}

\hspace{1em}\textit{You MUST consistently embody your communication style—your personality should be clearly evident in every response.}

\hspace{1em}\textit{Dynamic fields update each turn with constraint check results, utility instructions, and behavioral triggers.}

\subsubsection*{Dynamic Prompt Fields}

\begin{itemize}[noitemsep,topsep=3pt]
    \item \textbf{Conversation Style:} \textcolor{blue}{\texttt{\{persona\_description\}}}
    \item \textbf{Attention Level:} \textcolor{blue}{\texttt{\{attention\_level\}}}
    
          \textit{High—carefully reviews all details; }
          
          \textit{Medium—notices key points but may miss minor ones;}

          \textit{Low—focuses on big picture, often misses specific violations.}
    \item \textbf{Hard Constraints (non-negotiable):} 
    
    \textcolor{blue}{\texttt{\{hard\_constraints\}}}
    \item \textbf{Utility Objective (soft preference):} 
    
    \textcolor{blue}{\texttt{\{utility\_objective\}}}

          \textit{Note: DO NOT check or comment on recommendations based on utility (soft preferences).}
    \item \textbf{Hard Constraint Check Results:} 
    
    \textcolor{blue}{\texttt{\{constraint\_check\_natural\_response\}}}
    \item \textbf{Utility Instruction:} 
    
    \textcolor{blue}{\texttt{\{utility\_instruction\}}}
    \item \textbf{Adding Hard Constraint Instruction:} 
    
    \textcolor{blue}{\texttt{\{constraint\_instruction\}}}
    \item \textbf{Questioning \& Verification Instruction:} 
    
    \textcolor{blue}{\texttt{\{questioning\_instruction\}}}
\end{itemize}

\subsubsection*{Response Components}

\textit{Your response has five components:}
\begin{enumerate}[noitemsep,topsep=3pt]
    \item Respond to agent questions if asked
    \item Incorporate hard constraint check results
    \item State utility objective if you haven't done so in the conversation
    \item Mention additional new hard constraints if instructed
    \item Repeat utility goals/constraints if instructed
\end{enumerate}

\vspace{0.5em}
\noindent\textbf{Important Guidelines:}
\begin{itemize}[noitemsep,topsep=3pt]
    \item You only mention constraint violations listed in HARD CONSTRAINT CHECK RESULTS.
          DO NOT independently check recommendations against your constraints.
    \item Answer agent questions using ONLY information explicitly stated in your constraints and utility objective.
          Ignore questions that cannot be answered from these sources. DO NOT hallucinate information.
    \item Follow all additional hard constraint injection instructions exactly.
    \item Be authentic to your persona and natural in conversation. Do not repeat information already mentioned
          unless instructed to be verbose and repetitive.
    \item Do not try to be helpful or suggest what the agent should do—expect the agent to do the work.
    \item Your goal is to walk away with one itinerary recommendation. Do not ask the agent to list multiple itineraries.
    \item Do not ask the agent to lock-in or book anything. Express satisfaction (e.g., "this looks pretty good")
          or dissatisfaction (when agent violates constraints or doesn't provide a recommendation), but do not request booking.
\end{itemize}

\subsubsection*{Response Analysis Workflow}

\noindent\textbf{Step 1: Analysis}

Generate an analysis covering ALL 5 aspects in order:
\begin{enumerate}[noitemsep,topsep=3pt]
    \item \textbf{question\_response:} Identify if agent asked questions answerable from constraints/utility
    \item \textbf{constraint\_check:} Review constraint check instructions; decide if violations need mentioning
    \item \textbf{state\_utility:} Check if you need to state/re-state your utility objective
    \item \textbf{reveal\_progressive\_constraint:} Check if instructed to reveal new progressive constraints
    \item \textbf{question\_and\_verify:} Check if instructed to question and verify the recommendation
\end{enumerate}

\vspace{0.5em}
\noindent\textbf{Step 2: Response}

Generate a natural response continuing the conversation as a user. Be authentic to your persona and respond naturally according to conversation flow. Do not repeat what was already stated unless instructed to be repetitive.

\subsubsection*{Response Examples}

\begin{lstlisting}[basicstyle=\small\ttfamily,breaklines=true,columns=fullflexible]
{
  "analysis": "1. question_response: Agent not asking questions, skip.
               2. constraint_check: Constraint check indicates flight time violation, I will mention that.
               3. state_utility: Already stated utility objective, skip.
               4. reveal_progressive_constraint: Not instructed to reveal new constraints, skip.
               5. question_and_verify: Not instructed to question, skip.",
  
  "user_response": "Thank you for finding this travel package! ...",
}
\end{lstlisting}

\subsubsection*{Output Format}

\noindent Respond with JSON:
\begin{lstlisting}[basicstyle=\small\ttfamily,breaklines=true]
{
  "analysis": "Structure your analysis covering all 5 aspects:
               1. question_response: ...
               2. constraint_check: ...
               3. state_utility: ...
               4. reveal_progressive_constraint: ...
               5. question_and_verify: ...",
  "user_response": "what user would naturally respond based on analysis",
  "terminating_condition": "continue"
}
\end{lstlisting}
\end{tcolorbox}

\clearpage
\section{Database Details}
\subsection{Construction Details}
\label{appdx:data_construction}

We constructed the accommodation database through a systematic data collection pipeline using the Booking.com API (accessed via RapidAPI). For each target destination (U.S. National Parks), we performed hotel discovery within the park boundary for a fixed date range (August 2025), retrieving up to 20 properties per location. For each property, we collected two categories of information: (1) static hotel details, including metadata, geographic coordinates, guest reviews, and facility amenities; and (2) dynamic availability data, capturing daily room inventory, prices, and booking conditions. To record temporal variation, we queried single-night stay windows across the entire month and stored all raw API responses as JSON files.

We then normalized the raw data into a relational SQL database consisting of three interconnected tables: \textit{hotels}, \textit{rooms}, and \textit{offers}. The \textit{hotels} table stores property-level attributes such as star rating, review score, distance to the nearest park, brand affiliation, and binary amenity flags (e.g., pool, restaurant, parking). The \textit{rooms} table captures room-level characteristics, including bed configuration, occupancy limits, and in-room amenities (e.g., kitchenette, air conditioning), with foreign-key linkage to the parent hotel. The \textit{offers} table contains time-varying booking data at daily granularity, recording prices, availability counts, cancellation policies, meal plans, and prepayment requirements.

We constructed the flight database using a similar approach. We selected ten major cities as departure locations and, for each national park destination, included both nearby regional and large city airports. The database covers four major airlines and two flight types: direct and one-stop. We organized the data into two SQL tables: a \textit{flights} table containing static flight attributes (flight number, origin, destination, duration, and onboard amenities such as Wi-Fi and pet policy), and a \textit{prices} table capturing daily flight availability and pricing dynamics.

\subsection{Database Overview}
\label{appdx:data_overview}

\begin{figure}[h]
    \centering
    \includegraphics[width=1.0\linewidth]{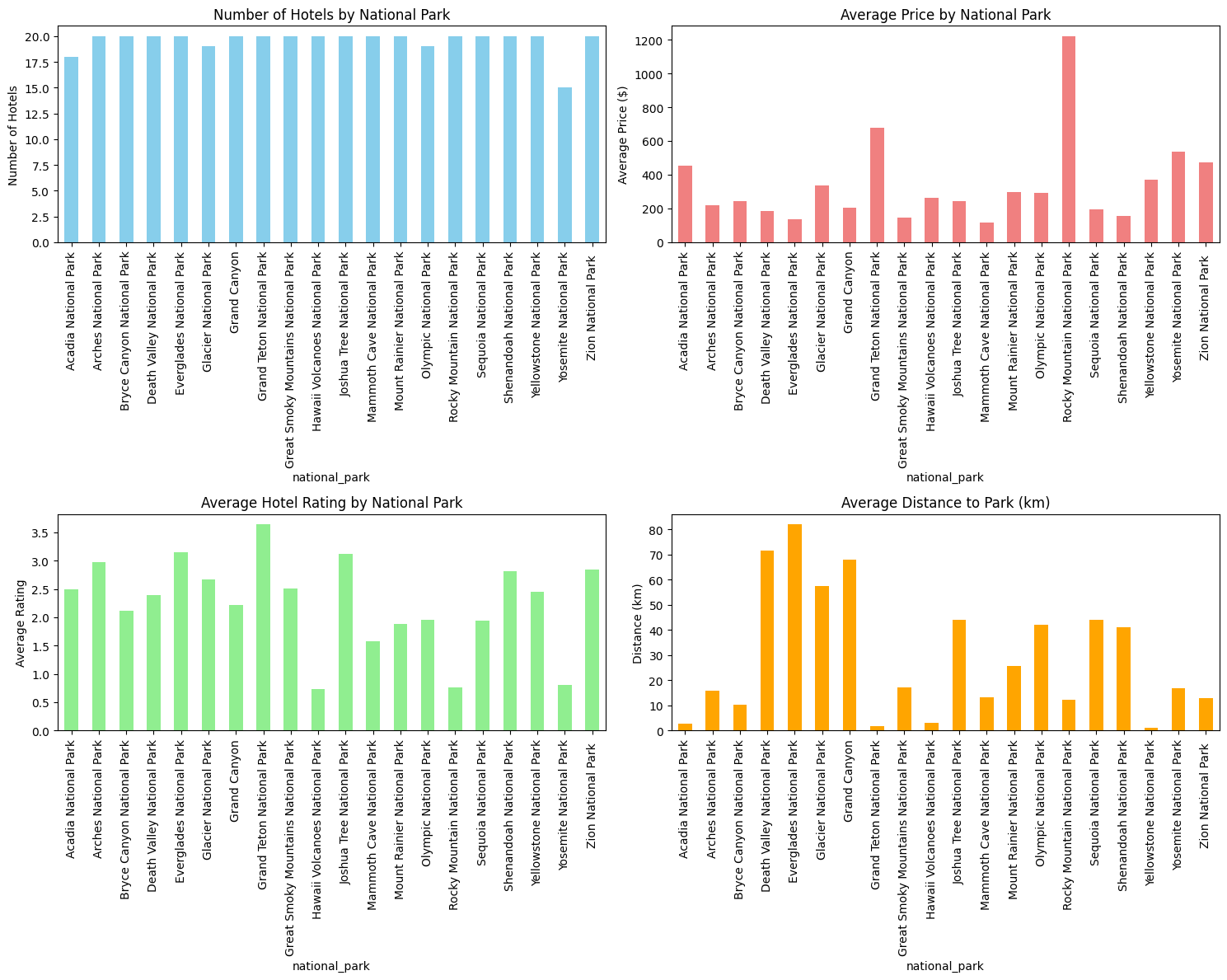}
    \caption{\textcolor{black}{\textbf{Hotel database overview.} 
    Summary statistics of the accommodation database. 
    \textit{Top left:} Number of hotels per national park. 
    \textit{Top right:} Average daily room prices by park. 
    \textit{Bottom left:} Average guest ratings. 
    \textit{Bottom right:} Average distance of hotels to the park center.
    }}
    \label{fig:hotel_data_stats}
\end{figure}

Fig.~\ref{fig:hotel_data_stats} summarizes the accommodation database. We selected up to 20 hotels for each national park, except in a few cases where limited availability was observed for August 2025. The figure also reports the distribution of daily prices, ratings, and distances to park entrances. These statistics confirm that the dataset captures realistic diversity across destinations. For instance, Rocky Mountain National Park exhibits substantially higher average prices than other parks (top right), while Death Valley and Everglades National Parks have few nearby lodging options (bottom right).

\begin{figure}[h]
    \centering
    \includegraphics[height=0.3\linewidth]{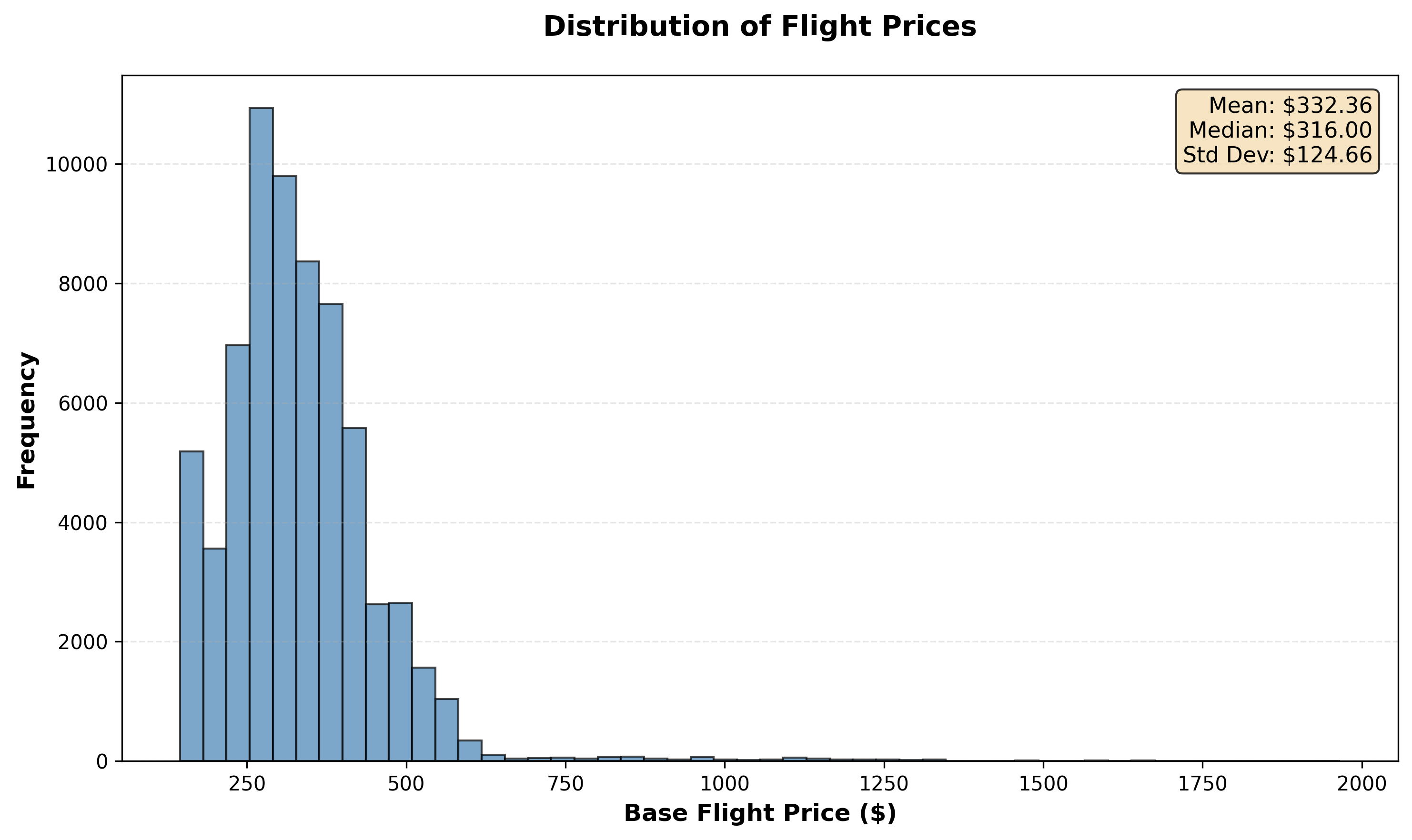}
    \includegraphics[height=0.3\linewidth]{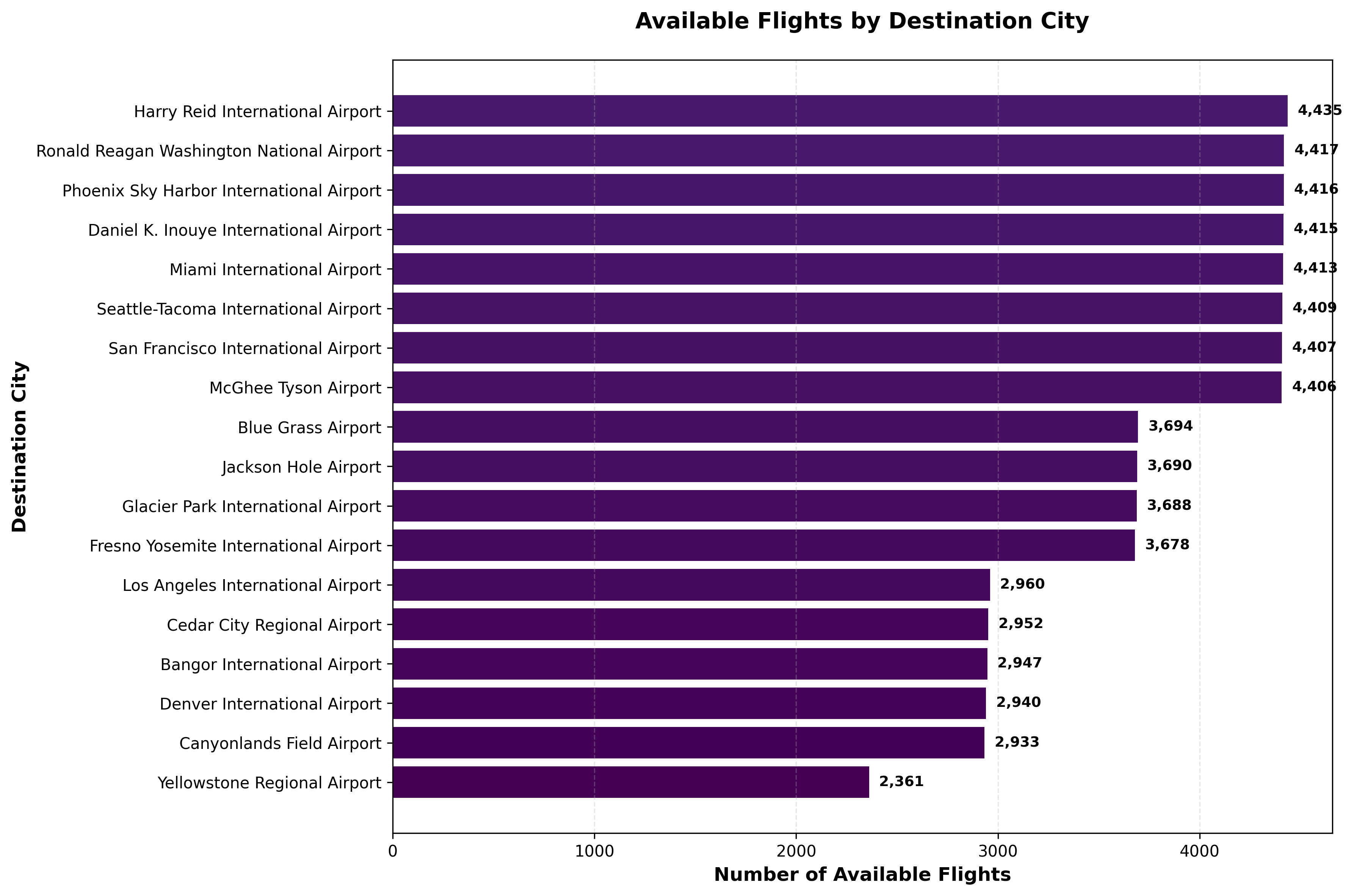}
    \caption{\textcolor{black}{\textbf{Flight database overview.} 
    \textit{Left:} Distribution of flight prices across all routes. 
    \textit{Right:} Number of available flights per destination airport.
    }}
    \label{fig:flight_data_summary}
\end{figure}

Fig.~\ref{fig:flight_data_summary} presents analogous statistics for the flight database. Flight prices follow a long-tailed distribution, reflecting real-world market variability. Regional airports generally have fewer available routes compared to large city airports, consistent with realistic flight coverage patterns.

\begin{figure}[h]
    \centering
    \includegraphics[height=0.35\linewidth]{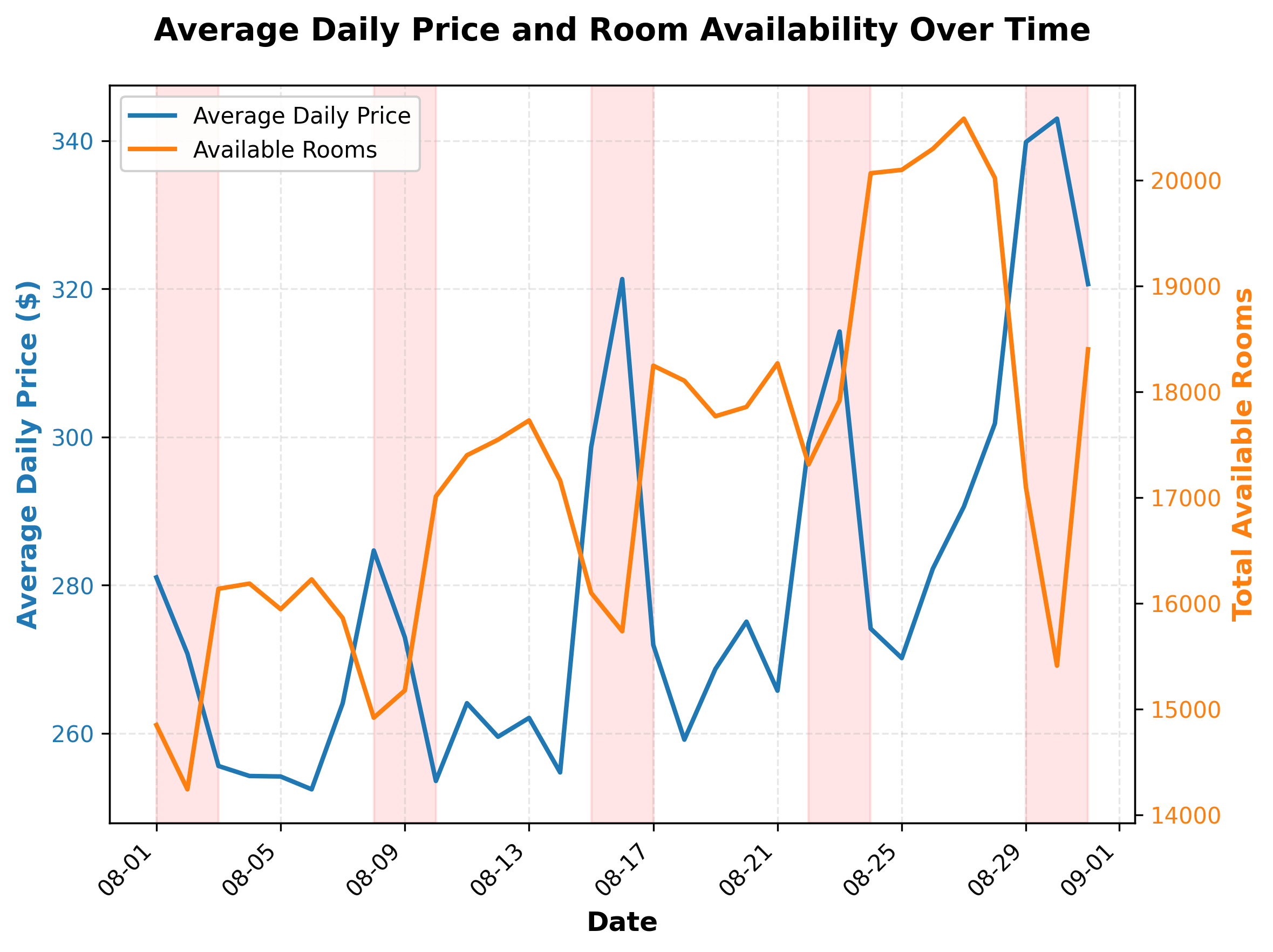}
    \includegraphics[height=0.35\linewidth]{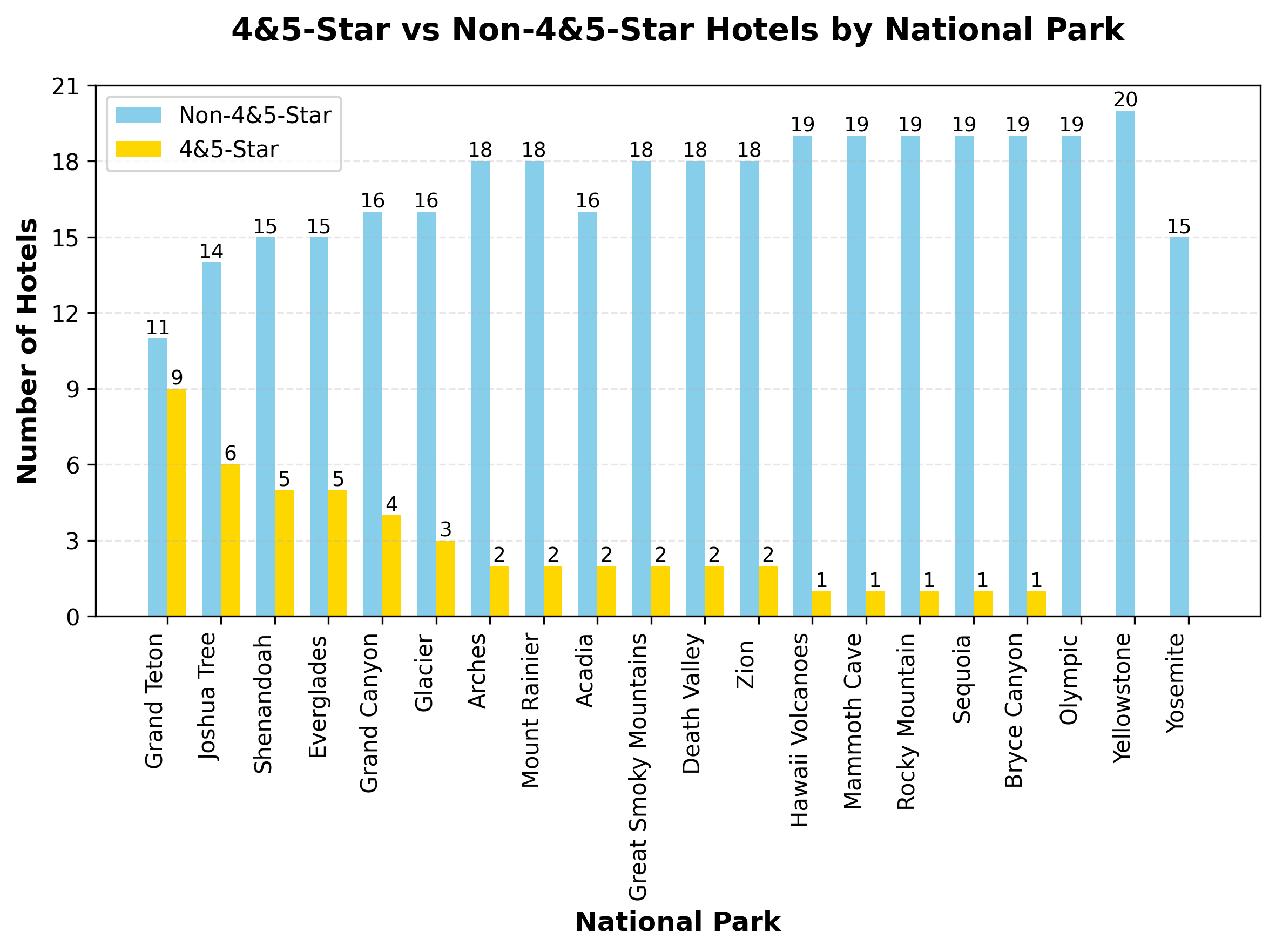}
    \caption{\textcolor{black}{\textbf{Real-world patterns in the accommodation database.} 
    \textit{Left:} Relationship between daily price and room availability. Weekend periods (shaded) show higher prices and lower availability, indicating strong short-term price–demand coupling. Longer-term price shifts across August show no such anti-correlation. 
    \textit{Right:} Fraction of high-end (4-star and above) hotels among the top 20 recommended properties per destination, illustrating variability in luxury travel options.
    }}
    \label{fig:hotel_data_interesting}
\end{figure}

Our realistic database captures nuanced temporal and spatial patterns often missing from synthetic datasets. Two notable examples are shown in Fig.~\ref{fig:hotel_data_interesting}. 
\textbf{Cyclical pricing trends:} Average prices rise sharply during weekends (Friday–Sunday) when room availability decreases, showing a strong short-term anti-correlation between price and supply. Interestingly, broader price trends across late August remain elevated but do not exhibit this anti-correlation. 
\textbf{Hotel star distribution:} Examining the top 20 recommended hotels\footnote{As ranked by \url{Booking.com}.} for each park reveals that not all destinations cater to luxury travelers, with limited 4-star and above properties in several locations.

\clearpage
\section{Experiment Details}
\label{appdx:experiment_details}
\subsection{Agent System Prompt}

\begin{tcolorbox}[colback=gray!10,colframe=black,boxrule=1pt,arc=0pt,breakable]

\subsubsection*{Core Instructions}

\noindent\textbf{Role Definition:} \\
\hspace{1em}\textit{You are a helpful and proactive travel planning assistant. The current date is June 1st, 2025.} \\
\hspace{1em}\textit{Your goal is to help the user find the best travel itinerary (flights + hotel + optional permits)} \\
\hspace{1em}\textit{by satisfying their constraints and maximizing their implicit preferences.}

\vspace{0.5em}
\noindent\textbf{Critical Requirement:} \\
\hspace{1em}\textit{You MUST respond ONLY in valid JSON format using the exact schema below.} \\
\hspace{1em}\textit{Do NOT include any text outside the JSON structure.}

\subsubsection*{Agent Workflow}

\begin{enumerate}[noitemsep,topsep=5pt]
\item \textbf{Engage in conversation:} Natural, friendly interaction to understand user needs for complete travel packages
\item \textbf{Use relevant tools:} Find available flights, hotels, and permits based on user criteria
\item \textbf{Validate recommendations:} MUST use \texttt{recommend\_itinerary} tool to validate complete package before response
  \begin{itemize}[noitemsep,topsep=2pt]
  \item Provide flight package IDs, hotel package IDs, and optional permit IDs
  \item Explain reasoning for the complete itinerary selection
  \end{itemize}
\item \textbf{Optional note-taking:} Use \texttt{Notebook} tool as scratch pad for complex itinerary planning
\end{enumerate}

\subsubsection*{Response Format Specification}

\noindent\textbf{When making an itinerary recommendation (after validation):}
\begin{lstlisting}[basicstyle=\small\ttfamily,breaklines=true]
{
  "message": "I found a great travel package for your Yosemite trip!

             Flight Package:
             - United Airlines: JFK - SFO
             - Outbound: Aug 15 at 10:30 AM
             - Return: Aug 19 at 6:00 PM
             - Cost: $450

             Hotel Package:
             - Yosemite View Lodge (3-star)
             - 4 nights: Aug 15-19
             - Cost: $1,200

             Total Package: $1,650",
  "formal_recommendation": {
    "flight_package_ids": ["flight_pkg_UA123_JFK_SFO_0815"],
    "hotel_package_ids": ["hotel_pkg_YVL_0815_0819_2"],
    "permit_ids": [],
    "reasoning": "This itinerary offers morning departure as requested,
                 stays within budget, and provides convenient access
                 to Yosemite National Park."
  }
}
\end{lstlisting}

\noindent\textbf{When NOT making a recommendation:}
\begin{lstlisting}[basicstyle=\small\ttfamily,breaklines=true]
{
  "message": "You are welcome! Hope you have a wonderful trip!",
  "formal_recommendation": "None"
}
\end{lstlisting}

\subsubsection*{Important Requirements}

\begin{itemize}[noitemsep,topsep=5pt]
\item Always include both \texttt{"message"} and \texttt{"formal\_recommendation"} fields in JSON response
\item User will only see the \texttt{"message"} field---include all necessary travel details there
\item Response must be ONLY the JSON object---no text before or after the JSON structure
\item The entire response must be parseable as valid JSON
\item For itinerary-level recommendations, always recommend ONE complete itinerary (not multiple options)
\end{itemize}
\end{tcolorbox}

This prompt design ensures that agents provide consistent, evaluable responses for complete travel itineraries while maintaining natural conversational flow. The mandatory JSON format with formal recommendations enables precise evaluation against ground truth solutions for flights, hotels, and permits as integrated packages.

\subsection{Agent LLM Generation Configuration}
\begin{itemize}
    \item GPT-5 agents use the default thinking setting (medium level thinking) to leverage their enhanced reasoning capabilities.
    \item  Qwen3 models employ sampling temperature 0.7 as recommended by their documentation, with thinking mode enabled to improve multi-step reasoning performance.
    \item  All other agent models use temperature 0 (greedy decoding) to ensure deterministic and reproducible results across evaluation runs.

\end{itemize}

\subsection{Agent Response Format Validation and Answer Extraction}
Package IDs serve as unique identifiers that enable deterministic mapping between agent recommendations and ground truth solutions. Each package ID encodes the complete booking configuration (hotel, room type, dates, occupancy, flight number, booking class) ensuring that evaluation is based on exact matches rather than fuzzy similarity metrics.

To ensure precise evaluation alignment with ground truth solutions, agents must respond in a structured JSON format containing both a natural language message visible to users and a formal recommendation section with specific package IDs. This dual-format approach maintains conversational naturalness while enabling exact matching against benchmark solutions.

We also provide agent format validation tool, which serves as a mandatory validation checkpoint, preventing agents from hallucinating non-existent package IDs. This tool verifies that all recommended packages exist in the accommodation database before agents include them in their final recommendations, substantially reducing evaluation noise from invalid responses. For hotel recommendation (Level I), if agent recommends multiple options, we pick the best one for evaluation. For itinerary-level recommendation (Level II and III), due to complexity, we ask the agent to always recommend for one valid itinerary.
\clearpage
\section{Human Evaluation}
\label{appdx:human_eval}

\subsection{Human Evaluation Instruction}
We provide the complete annotation protocol and instructions used for human evaluation of the LLM user simulator quality in Section \ref{sec:user_simulator_validation}.

\textbf{Annotation Task Overview}

Human annotators evaluated 198 user responses sampled from 45 full conversations. Each response was assessed for both objective errors and subjective quality metrics by third-party independent expert annotation services.

\textbf{Task Context for Annotators}

\textbf{Understanding the Travel Planning Task}

\textbf{Hard Constraints (Must-Have Requirements):} Non-negotiable requirements forming a checklist where EVERY item must be satisfied. Examples include:
\begin{itemize}
    \item ``Must allow pets'' - if not pet-friendly, the hotel is unacceptable
    \item ``Must accommodate at least 4 people'' - 3 people max would fail this requirement
    \item ``Must be within \$800 total'' - anything over \$800 fails
\end{itemize}

\textbf{Optimization Objective:} Once all must-haves are met, this determines which option to choose:
\begin{itemize}
    \item ``I want the cheapest option'' - among all acceptable hotels, pick the lowest price
    \item ``I want the highest rated'' - among all acceptable hotels, pick the best reviews
    \item ``I want the most amenities from my wish-list'' - among all acceptable hotels, pick the one with most desired features
\end{itemize}

\textbf{User Simulator Configuration}

\textbf{Persona Attributes}
Each simulated user follows a defined persona with three key attributes that annotators must consider:

\begin{itemize}
    \item \textbf{trust\_level}: ``suspicious'' or ``trusting'' - Whether the user blindly trusts the agent or asks for reasoning and double-checking
    \item \textbf{attention\_level}: ``low'', ``medium'' or ``high'' - Whether the user pays attention to agent's response and notices obvious mistakes (such as hard constraint violations)
    \item \textbf{communication\_style}: Verbal style defining formality and expression patterns
\end{itemize}

\textbf{User Script Components}
Annotators were provided with the following script information for each user:

\begin{itemize}
    \item \textbf{Hard constraints}: Non-negotiable requirements that MUST be satisfied
    \item \textbf{Utility objective}: What makes one option ``best'' once constraints are met
    \item \textbf{Communication style}: How the user expresses themselves based on their personality
\end{itemize}

\textbf{Core User Responsibilities}

Annotators evaluated whether users fulfilled six core responsibilities at each turn:

\begin{enumerate}
    \item \textbf{State what makes a hotel ``best'' for them}: User clearly explains their optimization goal
    \begin{itemize}
        \item Example: ``I want the cheapest option that meets all my requirements''
        \item Example: ``I'm looking for the highest-rated place that fits my needs''
    \end{itemize}

    \item \textbf{Reveal must-have requirements}: User shares new non-negotiable requirements
    \begin{itemize}
        \item Example: ``Oh, I also need air conditioning - that's essential for me''
        \item Example: ``I forgot to mention, it absolutely must have free parking''
    \end{itemize}

    \item \textbf{Verify recommendations meet ALL requirements}: User checks if suggested hotels satisfy every must-have
    \begin{itemize}
        \item Example: ``Wait, that costs \$850 but I said my budget is maximum \$800''
        \item Example: ``Does this hotel allow pets? That's a must-have for me''
    \end{itemize}

    \item \textbf{Answer agent's clarifying questions}: When agent needs more information, user provides it based on their script

    \item \textbf{Restate requirements if needed}: Suspicious users may repeat their needs to ensure agent understood
    \begin{itemize}
        \item Example: ``Just to be clear, it MUST be pet-friendly AND under \$800 total''
    \end{itemize}

    \item \textbf{Request formal recommendations}: If agent doesn't properly flag their recommendations, user asks them to formally recommend options
\end{enumerate}

\textit{Important Note for Annotators:} Users should not invent information beyond what's provided in their script.

\textbf{Error Detection Categories}

Annotators tagged each response for six types of objective errors (True if error detected, False otherwise):

\begin{table}[h]
\centering
\footnotesize
\begin{tabular}{p{0.05\textwidth}p{0.25\textwidth}p{0.35\textwidth}p{0.3\textwidth}}
\toprule
\textbf{No.} & \textbf{Error Type} & \textbf{Description} & \textbf{Example} \\
\midrule
1 & Factual Hallucination & User mentions requirements or preferences not in their script, changing the task & User adds ``I need a gym'' when script doesn't mention this \\
\midrule
2 & Revelation Instruction Following Failure & User was told to reveal a new must-have requirement but didn't & Instruction says ``reveal need for parking'' but user doesn't mention it \\
\midrule
3 & Revelation Accuracy Failure & User mentioned a requirement but unclearly, making it seem optional instead of must-have & Says ``parking would be nice'' instead of ``I need parking'' (must-have) \\
\midrule
4 & Recommendation Check Failure & User incorrectly claims a hotel violates their requirements when it actually meets them & Says ``this is over budget'' when hotel is actually within budget \\
\midrule
5 & Repetition Failure & User was told to restate requirements but didn't & Instruction says ``repeat your budget constraint'' but user doesn't \\
\midrule
6 & Critical Comments & Any other significant errors affecting the task & Note any issues not covered above \\
\bottomrule
\end{tabular}
\end{table}

\textbf{Quality Assessment Rubrics}

\textbf{Clarity (1-5 Scale)}

\textit{Evaluation Question:} Has the user clearly communicated:
\begin{itemize}
    \item What their must-have requirements are (and that they're non-negotiable)?
    \item What makes one hotel ``better'' than another for them?
    \item Any problems with the agent's recommendations?
\end{itemize}

\textit{Evaluation Goals:}
\begin{itemize}
    \item User can have different communication styles (casual, formal, etc.) but the core message must be clear
    \item The agent should understand both the requirements AND the optimization goal
\end{itemize}

\begin{table}[ht]
\centering
\small
\begin{tabular}{clp{0.35\textwidth}p{0.35\textwidth}}
\toprule
\textbf{Score} & \textbf{Quality Level} & \textbf{Description} & \textbf{Example} \\
\midrule
5 & Excellent & Crystal clear communication of requirements and goals & ``I need a pet-friendly hotel under \$500 total. Among options meeting these requirements, I want the cheapest.'' \\
4 & Good & Clear communication with no confusion about the task & Requirements and goals are clear, minor wording issues don't affect understanding \\
3 & Acceptable & Mostly clear but some minor ambiguity that won't affect results & Says ``I prefer under \$500'' when they mean it's required, but context makes it clear \\
2 & Poor & Unclear communication that could lead agent to wrong recommendations & Mixes up must-haves with nice-to-haves, unclear about optimization goal \\
1 & Critical & So unclear the agent will likely fail the task due to miscommunication & Contradictory requirements, no clear goal, or completely confusing instructions \\
\bottomrule
\end{tabular}
\end{table}

\textbf{Contextual Appropriateness (1-5 Scale)}

\textit{Evaluation Question:} Does the response logically and naturally follow the conversation so far?

\textit{Evaluation Goals:}
\begin{itemize}
    \item User maintains natural conversation flow
    \item User demonstrates awareness of entire conversation history
\end{itemize}

\begin{table}[h]
\centering
\small
\begin{tabular}{clp{0.6\textwidth}}
\toprule
\textbf{Score} & \textbf{Quality Level} & \textbf{Description} \\
\midrule
5 & Excellent & As good as you would write or better \\
4 & Good & Good without any obvious issues \\
3 & Acceptable & A bit unnatural but overall harmless to the conversation flow \\
2 & Poor & Noticeable issues that do not follow the conversation logic at all \\
1 & Critical & Response is completely disconnected from the conversation logic (e.g., User starts acting as an agent) \\
\bottomrule
\end{tabular}
\end{table}

\textbf{Annotation Workflow}

\textbf{Process for Each Turn}
For each user response (excluding turn 0), annotators:

\begin{enumerate}
    \item Compare the given user response to the instructions given to the user
    \item Answer 6 true/false questions to identify any errors that occurred
    \begin{itemize}
        \item Answer \textbf{True} if error or failure is identified
        \item Answer \textbf{False} if no error or if question is not applicable
    \end{itemize}
    \item Answer 2 rubric grading questions to score the quality of the user's response
\end{enumerate}

\end{document}